\documentclass[10pt,twocolumn,letterpaper]{article}

\usepackage[pagenumbers]{cvpr} % To force page numbers, e.g. for an arXiv version

\usepackage{graphicx}
\usepackage{amsmath}
\usepackage{amssymb}
\usepackage{booktabs}
\usepackage[pagebackref,breaklinks,colorlinks]{hyperref}

\usepackage[capitalize]{cleveref}
\crefname{section}{Sec.}{Secs.}
\Crefname{section}{Section}{Sections}
\Crefname{table}{Table}{Tables}
\crefname{table}{Tab.}{Tabs.}

\def\cvprPaperID{11163} % *** Enter the CVPR Paper ID here
\def\confName{CVPR}
\def\confYear{2023}

\begin{document}

\title{Learning a 3D Morphable Face Reflectance Model from Low-cost Data}

\author{
Yuxuan Han\textsuperscript{1}
\and Zhibo Wang\textsuperscript{2}
\and Feng Xu\textsuperscript{1}
\smallskip
\and
\textsuperscript{1}School of Software and BNRist, Tsinghua University\qquad 
\textsuperscript{2}SenseTime Research
}

% \maketitle
\twocolumn[{%
\renewcommand\twocolumn[1][]{#1}%
\maketitle

\begin{center}
    \vspace{-10pt}
    \centering
    \captionsetup{type=figure}
    \includegraphics[width=\textwidth]{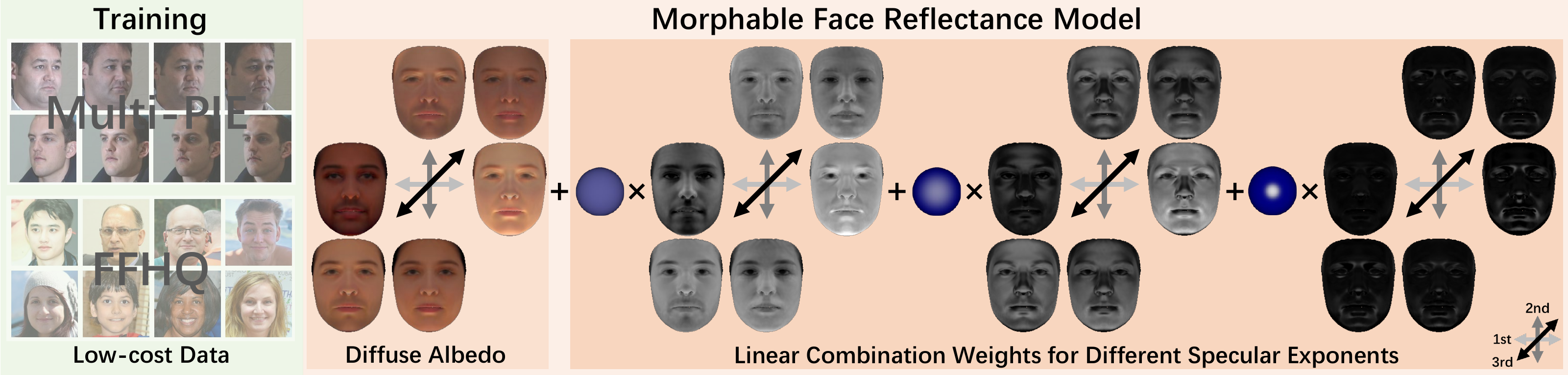}
    \vspace{-10pt}
    \captionof{figure}{We propose the first 3D morphable face reflectance model with spatially varying BRDF and a technique to train the model with low-cost publicly-available data. We represent face reflectance as a Lambertian BRDF combined with the linear combination of Blinn-Phong BRDFs with different predefined specular exponents. The reflectance parameters for each face vertex are the diffuse albedo and a set of weights. We show the first 3 principal components of diffuse albedo and spatially varying weights here in nonlinear sRGB space. }
    \label{Fig:teaser}
\end{center}%
}]

%%%%%%%%% ABSTRACT
\begin{abstract}
% \vspace{-22pt}
Modeling non-Lambertian effects such as facial specularity leads to a more realistic 3D Morphable Face Model. 
Existing works build parametric models for diffuse and specular albedo using Light Stage data. 
However, only diffuse and specular albedo cannot determine the full BRDF.
In addition, the requirement of Light Stage data is hard to fulfill for the research communities. 
This paper proposes the first 3D morphable face reflectance model with spatially varying BRDF using only low-cost publicly-available data.
We apply linear shiness weighting into parametric modeling to represent spatially varying specular intensity and shiness. 
Then an inverse rendering algorithm is developed to reconstruct the reflectance parameters from non-Light Stage data, which are used to train an initial morphable reflectance model. 
To enhance the model's generalization capability and expressive power, we further propose an update-by-reconstruction strategy to finetune it on an in-the-wild dataset. 
Experimental results show that our method obtains decent rendering results with plausible facial specularities.
Our code is released \href{https://yxuhan.github.io/ReflectanceMM/index.html}{\textcolor{magenta}{here}}.
\end{abstract}

%%%%%%%%% BODY TEXT
\section{Introduction}
\label{sec:intro}

3D Morphable Face Models (3DMM)~\cite{blanz1999morphable,egger20203d} have attracted much attention in the past two decades, as it provides a powerful and compact statistical prior of 3D face geometry and appearance with dense point-to-point correspondence to various downstream applications like face reconstruction~\cite{tewari2017mofa,deng2019accurate,tewari2018self,tewari2019fml,genova2018unsupervised}, rendering~\cite{deng2020disentangled,thies2016face2face,thies2020neural,tewari2020stylerig,zheng2022avatar}, and animation~\cite{chaudhuri2020personalized,feng2021learning,bai2021riggable,garrido2016reconstruction,yang2020facescape,cao20133d}.
Existing works~\cite{tewari2018self,tewari2019fml,tewari2021learning} have demonstrated promising results for improving the generalization capability and expressive power of 3DMM under the assumption that faces are Lambertian surfaces.
However, it is still challenging to model non-Lambertian effects such as facial specularity in 3DMM, which can lead to a more realistic face model.

A few recent works~\cite{smith2020morphable,Li2020LearningFO} involve non-Lambertian facial reflectance in the morphable face model. 
Using a Light Stage~\cite{debevec2000acquiring,ma2007rapid,ghosh2011multiview}, they capture diffuse and specular albedo maps of tens of participants. 
Then, they model the diffuse and specular albedo by training a PCA model~\cite{smith2020morphable} or a deep generative network~\cite{Li2020LearningFO} on the acquired data.
However, only the diffuse and specular albedo cannot determine the complete Bidirectional Reflectance Distribution Function (BRDF).
Thus, other works~\cite{dib2021practical,dib2021towards,dib2022s2f2} set the remaining reflectance parameters (\emph{e.g.} specular exponent for the Blinn-Phong BRDF~\cite{blinn1977models}, roughness for the Torrance-Sparrow BRDF~\cite{torrance1967theory}) of all face vertices to a reasonable value to characterize specular shiness and obtain the complete BRDF. 
As shown in Figure~\ref{Fig:exp_relight}, these spatially uniform parameters lead to unpleasing rendering results since face reflectance is inherently spatially varying~\cite{weyrich2006analysis}.
Besides, the requirement of Light Stage data is hard to fulfill since building a Light Stage is quite difficult, and no publicly available Light Stage dataset is sufficient to construct a 3DMM.

To overcome these limitations, we propose and train the first morphable face reflectance model with spatially varying BRDF from low-cost publicly-available data.
Inspired by previous works~\cite{pandey2021total,matusik2003data}, we represent face reflectance as a Lambertian BRDF combined with the linear combination of several Blinn-Phong BRDFs corresponding to different predefined specular exponents.
Thus, the reflectance parameters of each face vertex include an RGB color for the Lambertian BRDF and a set of weights for the Blinn-Phong BRDFs.
As illustrated in Figure~\ref{Fig:modulate}, our representation can naturally modulate specular intensity and shiness by adjusting the absolute and relative scales of the linear combination weights, respectively.
Compared to previous works~\cite{smith2020morphable,Li2020LearningFO} not modeling specular shiness, we define a complete BRDF by this representation in 3DMM.
Compared to the traditional Blinn-Phong BRDF that models specular intensity and shiness in a nonlinear formulation~\cite{blinn1977models}, our linear representation (Equation~\eqref{brdf}) is much easier to reconstruct the reflectance parameters from recorded images.
With this linear reflectance representation, we develop an inverse rendering approach to estimate the spatially varying reflectance parameters for the 128 selected identities in Multi-PIE~\cite{gross2010multi}, a public dataset with face images captured under controlled camera views and light directions.
Then, we learn a PCA model for the estimated reflectance parameters as our initial morphable face reflectance model.

Considering that the Multi-PIE dataset only contains 128 identities which is far from sufficient to capture the variability of human faces, we propose to finetune the initial model on a large-scale in-the-wild dataset, FFHQ~\cite{Karras2018ASG}, to improve its generalization capability and expressive power.
As the inputs are in-the-wild images with unknown lighting information, it is not easy to reconstruct accurate reflectance from them. 
Our key observation is that, on the one hand, we already have an initial parametric reflectance model that can better formulate the reflectance reconstruction from in-the-wild images. 
On the other hand, the reconstructed reflectance from in-the-wild data could provide feedback to enhance the face prior knowledge in our morphable reflectance model.
Based on this observation, we jointly reconstruct the face reflectance coefficients and update the parameters of our morphable face reflectance model (the mean and bases).
Another challenge here is to predict high-order spherical harmonics (SH) lighting~\cite{ramamoorthi2001efficient} for in-the-wild images, which is crucial for updating the high-frequency information of our non-Lambertian reflectance model~\cite{ramamoorthi2001signal}.
To solve this problem, we build another PCA model for real-world environment lighting in SH coefficients space, which largely reduces the searching space of the high-order SH coefficients.
During face reconstruction, we first predict the parameters of the PCA lighting model and then retrieve the high-order SH coefficients from it.
Finally, the in-the-wild images are well reconstructed with our parametric reflectance model, and the model itself is also updated gradually in this process to achieve high generalization capability and expressive power.

In summary, our contributions include:
\vspace{-8pt}
\begin{itemize}
   \item We propose the first 3D morphable face reflectance model with spatially varying BRDF and a technique to train the model with low-cost publicly-available data.\vspace{-8pt}
    \item We apply linear shiness weighting into parametric face modeling to represent spatially varying specular shiness and intensity and ease the process of reconstructing reflectance from images.\vspace{-8pt}
   \item We propose an update-by-reconstruction strategy to finetune our face reflectance model on an in-the-wild dataset, improving its generalization capability and expressive power.
\end{itemize}

\section{Related Work}
\paragraph{3D Morphable Face Model}
The origin 3DMM, proposed by Blanz and Vetter~\cite{blanz1999morphable}, learns a PCA model to represent 3D face shape and texture from 200 scans.
This seminal work has motivated substantial follow-ups in the past two decades~\cite{egger20203d}. 
Paysan et al.~\cite{Paysan2009A3F} propose the Basel Face Model (BFM), the first 3DMM available to the public. 
However, BFM and the original 3DMM can only model neutral faces. 
To handle expression variation, Li et al.~\cite{li2017learning} propose the FLAME model with additive expression bases trained from 4D scans.
Cao et al.~\cite{Cao2014FaceWarehouseA3} build a bilinear expression model from a database with multi-expression scans of the same person.
Another class of works attempts to better capture human face variation by scaling up the number of scans for 3DMM training. 
Dai et al.~\cite{Dai2017A3M} and Booth et al.~\cite{Booth2016A3M}  learn large-scale 3DMM from 1.2k and 10k subjects, respectively. 
However, all of these previous works approximate face as Lambertian surface and ignore the modeling of non-Lambertian reflectance. 
Recently, some works~\cite{smith2020morphable,Li2020LearningFO} build morphable models for diffuse and specular albedo using Light Stage scans~\cite{ma2007rapid,ghosh2011multiview,Stratou2011EffectOI}.
However, specular shiness is ignored in their model. 
In addition, their requirement on Light Stage scans is hard to fulfill for the research community.
% In our method, we learn a morphable face reflectance model with spatially varying BRDF using only low-cost publicly-available data.
Our method can represent both spatially varying specular intensity and shiness while only using low-cost publicly-available data as the training set.

More recently, some works~\cite{tewari2018self,tewari2019fml,tewari2021learning,tran2018nonlinear,tran2019towards} propose to learn 3DMM from large-scale 2D datasets by jointly performing face reconstruction and face model learning.
These methods can learn a 3DMM that generalizes well across the population. 
Tewari et al.~\cite{tewari2019fml} learn linear face shape and texture models from videos by designing novel loss functions to handle depth ambiguity. 
A follow-up work~\cite{tewari2021learning} learns a complete 3DMM, including shape, expression, and texture, from videos and neutral face images.
Tran et al.~\cite{tran2018nonlinear,tran2019towards} learn a non-linear 3DMM from 2D image collections, using deep neural networks to model face shape and texture.
Inspired by these works, we finetune our initial face reflectance model on an in-the-wild face image dataset to improve its generalization capability and expressive power.

\paragraph{Face Appearance Capture}
Existing methods for face appearance capture~\cite{Klehm2015RecentAI} fall into two categories: the image-based method and the model-based method.
The key idea of the image-based method is to capture a set of images to sample the light transport function, and then novel appearances can be obtained by linearly recombining these images.
To fulfill this, Debevec et al.~\cite{debevec2000acquiring} construct the Light Stage to capture the light transport function by programmatically activating One-Light-At-a-Time (OLAT).
% Rencent works~\cite{Sun2020LightSS,Zhang2020NeuralLT} propose learning-based methods to increase the Light Stage's sampling rate on the light transport function.
Using the captured OLAT data as the training set, Mallikarjun et al.~\cite{MallikarjunB2021MonocularRO} propose a learning-based method to infer the whole light transport function from a monocular input face image, and Kumar et al.~\cite{Kumar2010MorphableRF} build a statistical model for the light transport function at a fixed frontal viewpoint.
By directly modeling the light transport function, these image-based methods can represent specularities, sub-surface scattering, and other high-order effects caused by the complex interaction between light and face surface.
However, they cannot export geometry or material assets for further usages like material editing or animation. 

Model-based methods capture the parameters of a reflectance model and utilize the rendering equation~\cite{kajiya1986rendering} to synthesize novel appearances.
Previous works~\cite{ma2007rapid,ghosh2011multiview,Stratou2011EffectOI} use polarised illumination to directly capture the diffuse and specular albedo of human face.
Using the captured data from~\cite{ma2007rapid,ghosh2011multiview,Stratou2011EffectOI}, recent works train a neural network to map a monocular face image into its diffuse and specular albedo map~\cite{Yamaguchi2018HighfidelityFR,Lattas2020AvatarMeRR} or build a morphable model for this maps~\cite {smith2020morphable,Li2020LearningFO}.
Another class of works adopts an inverse rendering framework to estimate the face reflectance parameters from images. 
Weyrich et al.~\cite{weyrich2006analysis} develop a novel reflectance model for face and estimate its parameter from dense OLAT images captured by a Light Stage~\cite{debevec2000acquiring}. 
Riviere et al.~\cite{Riviere2020SingleshotHF} leverage images captured by a lightweight single-shot system for inverse rendering.
In our method, we estimate the reflectance parameters for each identity in the Multi-PIE dataset from the provided OLAT images via an inverse rendering framework and use these parameters to build an initial morphable face reflectance model.

More recently, some works attempt to capture face appearance using a low-cost setup, such as a single selfie video of the subject rotating under the sun~\cite{Wang2022SunStagePR}, recorded video of the subject illuminated by a desktop monitor~\cite{Sengupta2021ALS}, and co-located captured sequence~\cite{Azinovic2022HighResFA,Sevastopolsky2020Relightable3H}.
However, all of these previous works are person-specific.
Our goal is to build a generic morphable reflectance model using low-cost data.

\section{Method}
In this Section, we first introduce the representation of our morphable face reflectance model (Section~\ref{sec:method:representation}), then propose a method to learn this model from low-cost publicly-available data.
Specifically, we first learn an initial model from the Multi-PIE dataset (Section~\ref{sec:method:init}) and then finetune it on the FFHQ dataset to improve its generalization capability and expressive power (Section~\ref{sec:method:finetune}).

\begin{figure}[t]
    \centering
    \includegraphics[width=0.45\textwidth]{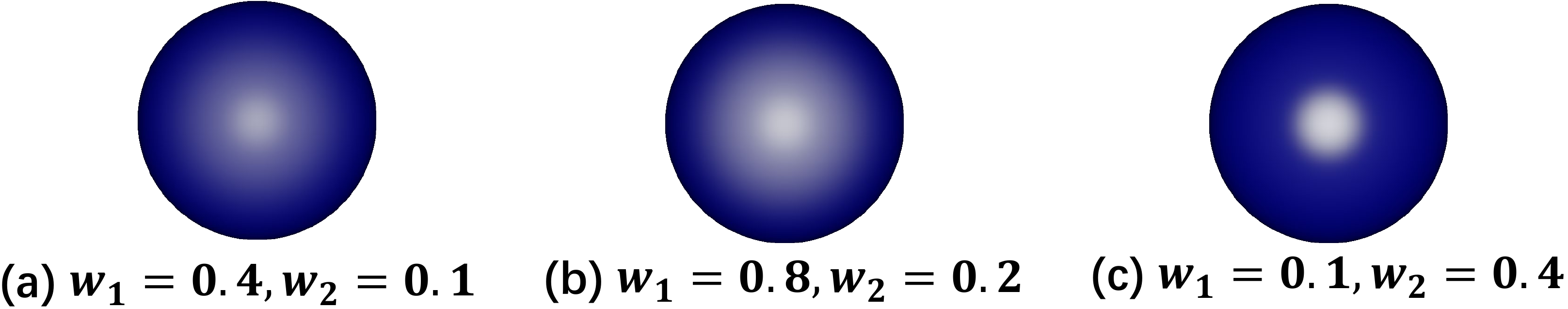}
    \vspace{-5pt}
    \caption{Our reflectance representation can naturally modulate specular intensity and shiness by adjusting the absolute or relative scales of the linear combination weights. Here we show an example using the linear combination of 2 Blinn-Phong BRDFs with specular exponents $p_1\!\!=\!\!8$ and $p_2\!\!=\!\!64$, respectively. Note the changes in specular intensity and shiness under different linear combination weights $w_1$ and $w_2$. }
    \label{Fig:modulate}
\end{figure}

\subsection{Morphable Face Reflectance Model}
\label{sec:method:representation}
Our goal is to design a morphable model to represent the spatially varying BRDF of the human face across the population. 
To this end, we build our model upon the BFM09~\cite{Paysan2009A3F} geometry model and assign the spatially varying reflectance parameters to its vertices. 
We employ a linear model for the reflectance parameters of the human face: 
\begin{equation}
   R = \bar{R} + {\rm M_R}\cdot \beta
   \label{morphable_reflectance_model}
\end{equation}
Here, $\bar{R}\in\mathbb{R}^{kV}$ and ${\rm {M_R}}\in\mathbb{R}^{kV\times N_R}$ are the mean and bases of face reflectance parameters, respectively; $N_R$ is the number of bases; $k$ is the number of reflectance model parameters for each vertex; $V$ is the number of vertices of the BFM09 geometry model; $\beta\in\mathbb{R}^{N_R}$ is the morphable model coefficients. 
Note that previous works~\cite{gerig2018morphable,tewari2019fml,tewari2021learning} represent face reflectance as the Lambertian BRDF. 
So in their scenario $k=3$ to represent the RGB diffuse color.  

Next, we first introduce our reflectance representation. 
Then, we illustrate our efficient shading technique for this representation under directional or environmental illumination, which can accelerate the model learning process detailed in later Sections. 

\vspace{-5pt}
\paragraph{Reflectance Representation}
To model non-Lambertian effects such as facial specularity, we incorporate a diffuse term and a specular term in our face reflectance representation $f_r$.
We instantiate them as the Lambertian BRDF and the linear combination of several Blinn-Phong BRDFs~\cite{blinn1977models} with different predefined specular exponents, respectively:
% Thus, we have:
\begin{equation}
    f_r(\textbf{l},\textbf{v},\textbf{n}) = \frac{c}{\pi} + \sum_{i=1}^{k_{bp}}w_i\cdot f_i\cdot \frac{\langle\textbf{h},\textbf{n}\rangle^{p_i}}{\langle\textbf{l},\textbf{n}\rangle}
    \label{brdf}
\end{equation}
Here, $\textbf{l}$, $\textbf{v}$, and $\textbf{n}$ indicate the incident light direction, viewing direction, and normal direction, respectively; $c$ is the RGB diffuse color; $w_i$ are the linear combination weights; $p_i$ are the predefined specular exponents; $k_{bp}$ is the number of Blinn-Phong BRDFs; $f_i=\frac{p_i+2}{4\pi\cdot(2-2^{-\frac{p_i}{2}})}$ are the energy normalization factor~\cite{akenine2019real} such that the corresponding Blinn-Phong lobe integrates to 1; $\langle\cdot,\cdot\rangle$ is the clamped cosine function; $\textbf{h}=\frac{\textbf{v}+\textbf{l}}{||\textbf{v}+\textbf{l}||_2}$ is the half vector~\cite{akenine2019real}.

In our scenario, the reflectance parameters for each face vertex are the diffuse color $c$ and $k_{bp}$ linear combination weights $w_i$.
Thus, each face vertex has $k=k_{bp}+3$ reflectance parameters attached to it.
% We assume a white specular albedo considering that skin is non-metallic material.
Note that the specular exponents $p_i$ are predefined and shared by each face vertex; they are hyper-parameters in our model.
Our representation can naturally modulate the specular intensity and shiness.
As illustrated in Figure~\ref{Fig:modulate}, doubling all the weights would double the specular intensity, while adjusting the aspect ratio between weights would change the specular shiness.
Moreover, our reflectance representation generalizes to previous work~\cite{dib2021practical,dib2021towards,dib2022s2f2} with spatially varying specular albedo and a global specular exponent when $k_{bp}=1$.

\vspace{-5pt}
\paragraph{Efficient Shading}
For \emph{directional illumination}, by denoting the incoming irradiance from direction $\textbf{l}$ as $E$, we can directly obtain the shading:
\begin{equation}
    s = E\cdot( \frac{c}{\pi}\cdot\langle\textbf{l},\textbf{n}\rangle + \sum_{i=1}^{k_{bp}}w_i\cdot f_i\cdot \langle\textbf{h},\textbf{n}\rangle^{p_i})
    \label{render_point_light}
\end{equation}

For \emph{environmental illumination}, we denote the incoming radiance from direction $\textbf{l}$ as $E(\textbf{l})$.
According to the rendering equation~\cite{kajiya1986rendering}, we have:
\begin{equation}
    s = \int_{\textbf{l}\in\Omega^+} E(\textbf{l})\cdot f_r(\textbf{l},\textbf{v},\textbf{n})\cdot \langle\textbf{l},\textbf{n}\rangle {\rm d}\textbf{l}
    \label{render_equation}
\end{equation}
Here, $\Omega^+$ is the upper hemisphere centered by $\textbf{n}$.
Substituting Equation~\eqref{brdf} into Equation~\eqref{render_equation}, we then separate the shading $s$ into a diffuse part $s_{d}$ and a specular part $s_{s}$:
\begin{align}
    s &= s_d + s_s, {\rm where} \\
    s_d &= \int_{\textbf{l}\in\Omega^+} \frac{c}{\pi}\cdot E(\textbf{l})\cdot  \langle\textbf{l},\textbf{n}\rangle {\rm d}\textbf{l} \\
    s_s &= \int_{\textbf{l}\in\Omega^+} \sum_{i=1}^{k_{bp}}w_i\cdot f_i {\langle\textbf{h},\textbf{n}\rangle^{p_i}}\cdot E(\textbf{l}) {\rm d}\textbf{l}
\end{align}
We can efficiently compute $s_d$ and $s_s$ in the frequency space~\cite{ramamoorthi2001signal}:
\begin{align}
    \label{env_light_render_diff} s_d &= \frac{c}{\pi}\cdot \sum_{l=0}^L\sum_{m=-l}^{l}A_l\cdot K_{lm}\cdot Y_{lm}(\textbf{n}),\\
    \label{env_light_render_spec} s_s &= \sum_{i=1}^{k_{bp}}\sum_{l=0}^L\sum_{m=-l}^{l}w_i\cdot B_l\cdot K_{lm}\cdot Y_{lm}(\textbf{r}).
\end{align}
Here, $L$ is the SH order, $A_l$, $B_l^i$, and $K_{lm}$ are the SH coefficients of the Lambertian BRDF, Blinn-Phong BRDF with specular exponent $p_i$, and the environmental illumination, respectively; $Y_{lm}(\cdot)$ are the SH basis functions; $\textbf{r}=\frac{2(\textbf{n}\cdot\textbf{v})\textbf{n}-\textbf{v}}{||2(\textbf{n}\cdot\textbf{v})\textbf{n}-\textbf{v}||_2}$ is the specular reflect direction~\cite{akenine2019real}. 

\begin{figure*}[t]
    \centering
    \includegraphics[width=0.75\textwidth]{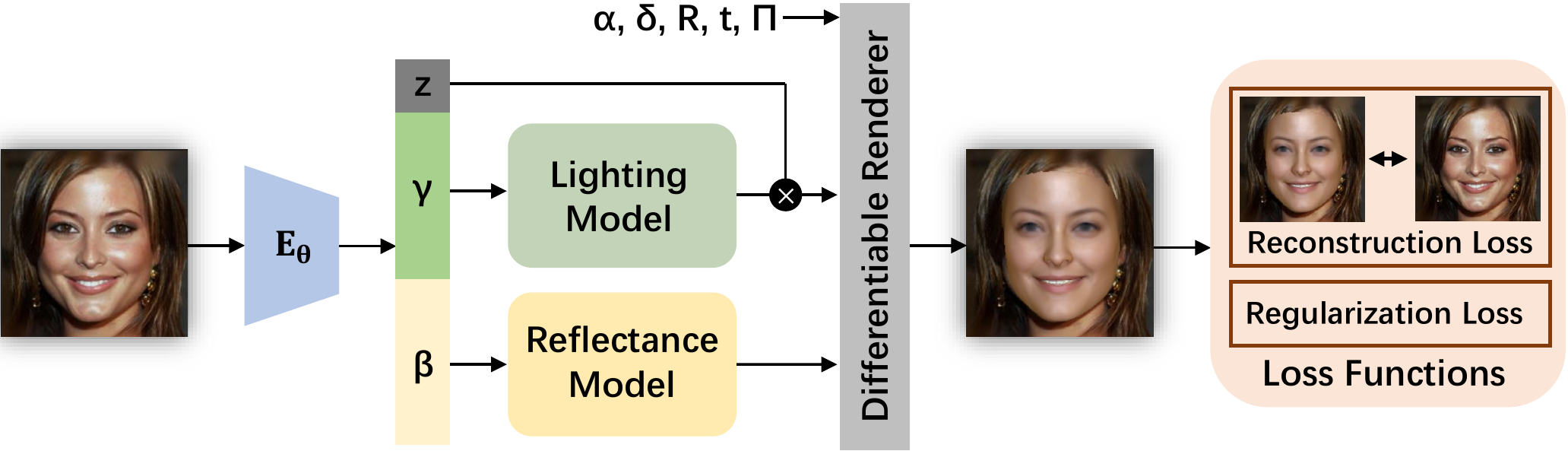}
    \caption{Model finetuning pipeline overview. Given a single input face image, we apply an encoder $E_{\theta}$ to estimate its lighting scale $z$, lighting coefficients $\gamma$, and reflectance coefficients $\beta$. Combined with the precompute geometry parameters $\alpha$, $\delta$, $\textbf{R}$, $\textbf{t}$, and $\Pi$, we can obtain the reconstructed face image via a differentiable renderer to compute self-supervised loss and jointly update $E_{\theta}$ and our reflectance model. }
    \label{Fig:framework}
\end{figure*}

\subsection{Initial Model Learning}
\label{sec:method:init}
In this part, we propose a method to learn the mean $\bar{R}$ and bases ${\rm M_R}$ of our morphable face reflectance model from the publicly-available Multi-PIE dataset~\cite{gross2010multi}.
Specifically, we first estimate the reflectance parameters for each identity in the dataset via inverse rendering, and then train a PCA model for them.

\vspace{-5pt}
\paragraph{Dataset Preprocessing}
The Multi-PIE dataset contains 337 identities captured under 15 viewpoints and 19 illuminations.
We first exclude objects with facial accessories or hair occlusions, resulting in 128 identities.
Then, we manually select 9 viewpoints and 12 illuminations (including 11 directional flash images and 1 room light image) with well color consistency to train our model.
By removing the room light effect in the flash images, we obtain 11 OLAT images per viewpoint.
We adopt a simple model-based approach to reconstruct the BFM09 geometry coefficients of each identity, camera parameters of each viewpoint, and the position of each flash simultaneously. 
See more implementation details in our \emph{Supplementary Material}.

\vspace{-5pt}
\paragraph{Refectance Parameter Estimation}
For a specific identity, we estimate all the reflectance parameters in UV space, including the diffuse color map $C\in\mathbb{R}^{3\times H\times W}$ and the linear combination weight map $W\in\mathbb{R}^{{k_{bp}}\times H\times W}$.
We unwarp all the OLAT images into UV space and denote the one captured under the $i$-th viewpoint and the $j$-th directional flash illumination as $I^{uv}_{ij}\in\mathbb{R}^{3\times H\times W}$.
From the reconstructed face geometry and scene information, we precompute the incident light direction $\textbf{l}^{uv}_{j}$ for each flash, view direction $\textbf{v}^{uv}_{i}$ for each camera, normal direction $\textbf{n}^{uv}$ for the face geometry, and shadow mask\footnote{We obtained shadow mask via ray tracing.} $M^{uv}_{ij}$ for each OLAT image in the UV space.
By predefining a reasonable incoming irradiance $E$ from the directional flash\footnote{There is an inevitably global scale between the reflectance parameters estimated by the inverse rendering method and the ground truth, if lighting unknown. See more theoretical analysis in~\cite{ramamoorthi2001signal}. }, we obtain the reconstructed OLAT image $\hat{I}^{uv}_{ij}$ using the efficient shading technique under directional illumination presented in Equation~\eqref{render_point_light}:
\begin{equation}
    \hat{I}^{uv}_{ij} = E\cdot(\frac{C}{\pi} \cdot \langle\textbf{l}^{uv}_{j},\textbf{n}^{uv}\rangle + \sum_{k=1}^{k_{bp}}W_k\cdot f_k\cdot \langle\textbf{h}^{uv}_{ij},\textbf{n}^{uv}\rangle^{p_k})
\end{equation}
Here, $\textbf{h}^{uv}_{ij}$ is the half vector UV map obtained by $\textbf{l}^{uv}_{j}$ and $\textbf{v}^{uv}_{i}$.
We optimize the diffuse color map $C$ and linear combination weights map $W$ with loss:
\begin{equation}
    \mathop{\arg\min}\limits_{C,W} \mathcal{L}_{recon} + w_{reg}\mathcal{L}_{reg}
\end{equation}
$\mathcal{L}_{recon}$ is the weighted L1 reconstruction loss:
\begin{equation}
    \mathcal{L}_{recon} = \sum_{i,j} \langle\textbf{l}^{uv}_{j},\textbf{n}^{uv}\rangle\cdot  ||M^{uv}_{ij}\cdot(\hat{I}^{uv}_{ij} - I^{uv}_{ij})||_1,
\end{equation}
$\mathcal{L}_{reg}$ is designed to restrict the reflectance parameters to be non-negative:
\begin{equation}
    \mathcal{L}_{reg} = -M_{C}\cdot C - M_{W}\cdot W
\end{equation}
Here, $M_{C}$ and $M_{W}$ are the masks that indicate negative values in $C$ and $W$, respectively.
During parameter estimation, we randomly horizontal flip $C$ and $W$ to introduce symmetric constraints~\cite{Wu_2020_CVPR} to the reflectance parameter map.

Compared to the traditional way that uses specular intensity and exponents to parameterize the Blinn-Phong BRDF, our linear representation is much easier for parameter estimation in practive.

\vspace{-5pt}
\paragraph{Model Learning}
With the estimated reflectance parameter maps for each identity, we can build our initial morphable face reflectance model. 
Similar to AlbedoMM~\cite{smith2020morphable}, we learn a PCA model only for the diffuse albedo.
Then, we transfer it to the specular weights by using the same linear combination of the training samples to form the bases. 
Thus, we can use the same coefficients $\beta$ for the diffuse and specular reflectance parameters as Equation~\eqref{morphable_reflectance_model} while keeping the orthonormality of the diffuse bases so that the user can use our diffuse model independently.

\subsection{Model Finetuning}
\label{sec:method:finetune}
To improve the generalization capability and expressive power of our initial morphable face reflecatance model, we finetune it on an in-the-wild face image dataset, FFHQ~\cite{Karras2018ASG}, by jointly doing face reconstruction and model updating.

\vspace{-5pt}
\paragraph{Dataset Preprocessing}
Before model finetuning, we use an off-the-shelf~\cite{deng2019accurate} method to estimate the BFM09 geometry coefficients and head pose for each image in the dataset.
To further improve the geometry reconstruction accuracy, we apply an offline optimization using the same loss functions as~\cite{deng2019accurate}. 
Finally, we obtain the shape coefficients $\alpha$, expression coefficients $\delta$\footnote{The expression bases are adapted from FaceWarehouse~\cite{Cao2014FaceWarehouseA3} since BFM09 does not model expression. See more details in ~\cite{deng2019accurate}.}, and head pose $\textbf{R},\textbf{t}$ for each image.
Similar to~\cite{deng2019accurate}, we use the perspective camera model with a reasonable predefined focal length to represent the 3D-2D projection $\Pi$.

\vspace{-5pt}
\paragraph{Network Architecture}
As illustrated in Figure~\ref{Fig:framework}, given a single face image $I$ as input, our face reconstruction network $E_\theta(\cdot)$ predicts the reflectance model coefficients $\beta$ and the SH lighting.
Combined with the geometry parameters $\alpha,\delta$, head pose $\textbf{R},\textbf{t}$, and the projection $\Pi$, we can obtain the reconstructed image $\hat{I}$ via a differentiable rasterizer~\cite{Ravi2020Accelerating3D,Laine2020ModularPF} using the efficient shading technique presented in Equation~\eqref{env_light_render_diff} and \eqref{env_light_render_spec}.

To update the high-frequency information in our non-Lambertian reflectance representation, we need to predict high-order SH lighting~\cite{ramamoorthi2001signal}.
We adopt 8-order SH lighting with 273 parameters in our method as ~\cite{dib2021towards,Li2014IntrinsicFI}.
However, if handled naively, the network cannot predict reasonable SH lighting due to the large searching space of high-order SH coefficients, as shown in Figure~\ref{Fig:exp_ablat}.
To constrain the searching space, we build a PCA model for the real-world environmental lighting in SH coefficient space inspired by~\cite{egger2018occlusion}.
Specifically, we utilize a real-world HDRI environment map dataset~\cite{Wang2020SingleIP} and apply rotation augmentation to it.
For each environment map, we compute its SH coefficients up to the 8-$th$ order.
We then divide them by the 0-$th$ order coefficient for normalization.
Note that each color channel is normalized independently.
Next, we learn a PCA model for these normalized SH coefficients.
We use the first $N_L$ bases for lighting prediction.

During finetuning, together with the reflectance model coefficients $\beta$, our network predicts $\gamma\in\mathbb{R}^{N_L}$ as the lighting PCA model coefficients and $z\in\mathbb{R}^{3}$ to represent the scale of 0-$th$ order SH coefficient for each color channel.
From this, we first use $\gamma$ to recover the SH coefficients from the PCA lighting model and then apply the predicted scale $z$ to them.
We adopt the ResNet-50~\cite{He2016DeepRL} architecture as the reconstruction network $E_\theta(\cdot)$ and modify the last fully-connected layer to $N_R+N_L+3$ neurons.
We adopt the Softplus activation for $z$ to ensure a non-negative prediction and linear activation for $\beta$ and $\gamma$.

\vspace{-5pt}
\paragraph{Loss Function}
In model finetuning, the learnable parameters are the morphable model parameters, including the mean $\bar{R}$ and bases ${\rm M_R}$, and face reconstruction network parameters $\theta$. 
We optimize them with the combination of a reconstruction loss $\mathcal{L}_{rec}$ and a regularization loss $\mathcal{L}_{reg}$:
\begin{equation}
    \mathop{\arg\min}\limits_{\bar{R},{\rm M_R},\theta} \mathcal{L}_{rec} + \mathcal{L}_{reg}
\end{equation}
$\mathcal{L}_{rec}$ is the combination of a L1 term $\mathcal{L}_{l1}$ and a perceptual term $\mathcal{L}_{per}$; see more details in our \emph{Supplementary Material}.
In our regularization loss $\mathcal{L}_{reg}$, we design $\mathcal{L}_{upd}$ to constrain the updating of our morphable reflectance model:
\begin{equation}
    \mathcal{L}_{upd} = ||\bar{R}-\bar{R}_0||_1 + ||\rm{M_R} - \rm{M_{R_0}}||_1 
\end{equation}
In addition, we adopt $\mathcal{L}_{light}$ to encourage monochromatic environment lighting as~\cite{DFRgithub} to resolve the color ambiguity between albedo and lighting and $\mathcal{L}_{coef}$ to constrain the predicted PCA coefficients $\beta$ and $\gamma$; see more details in our \emph{Supplementary Material}.

\begin{figure}[t]
    \centering
    \includegraphics[width=0.45\textwidth]{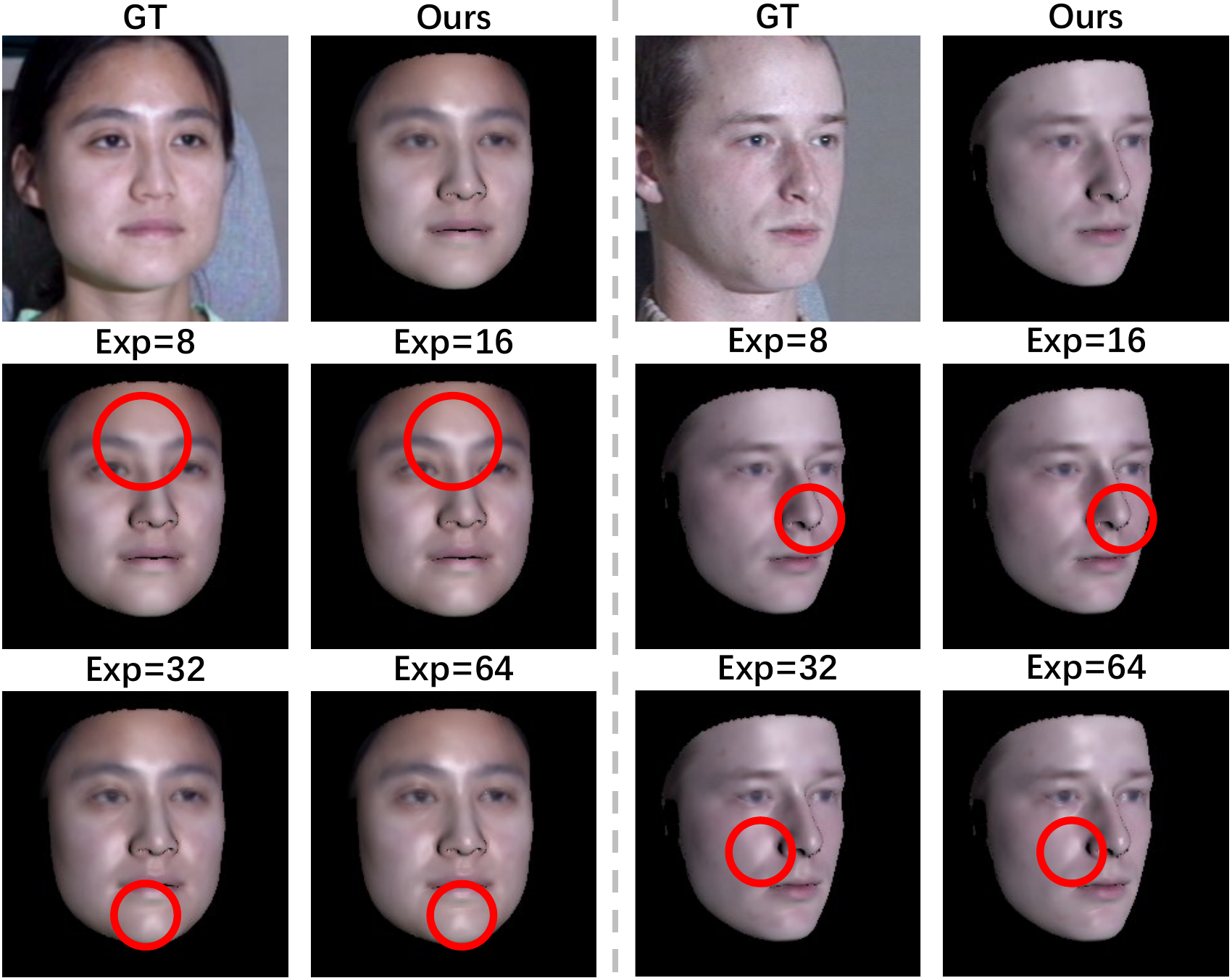}
    \vspace{-5pt}
    \caption{Qualitative comparison of face rendering results on the Multi-PIE dataset between our spatially varying reflectance representation and previous works with a global specular exponent. Here, \emph{Exp} stands for the specular exponent in the Blinn-Phong BRDF. A large global exponent (\emph{e.g.} 32, 64) leads to over-shiness artifacts around the cheek and chin, while a small one (\emph{e.g.} 8, 16) cannot represent specularities appear in the forehead and tip of the nose. Our representation can well model the spatially varying specular intensity and shiness on the face. }
    \label{Fig:exp_cmp_global_shiness}
\end{figure}

\section{Experiments}
\subsection{Implementation Details}
\paragraph{Initial Model Learning}
For reflectance parameter estimation, we set $w_{reg}\!\!=\!\!100$ and adopt Adam~\cite{Kingma2015AdamAM} optimizer to minimize the loss function, with learning rate 5e-3.
We use 3 Blinn-Phong BRDFs with specular exponents 1, 8, and 64 in our method, \emph{i.e.} $k_{bp}\!\!=\!\!3$.
Thus, there are 6 reflectance parameters for each face vertex.
We use the first 80 PCA bases in our initial model, \emph{i.e.} $N_R\!\!=\!\!80$.

\vspace{-5pt}
\paragraph{Model Finetuning}
For the lighting PCA model, we also use the first 80 PCA bases, \emph{i.e.} $N_L\!\!=\!\!80$.
The weights for $\mathcal{L}_{l1}$,$\mathcal{L}_{per},\mathcal{L}_{coef},\mathcal{L}_{upd},\mathcal{L}_{light}$ are set to $2, 0.1, 0.001, 10, 10$, respectively. 
We finetune our model on the FFHQ dataset~\cite{karras2020analyzing}, with 70000 high-fidelity single-view face images, and crop them to $224\!\!\times\!\!224$ when input to our reconstruction network $E_{\theta}$.
We first pretrain $E_{\theta}$ using $\mathcal{L}_{l1}$,$\mathcal{L}_{per}$,$\mathcal{L}_{coef}$ for 20 epochs to ensure it can output reasonable reflectance and lighting coefficients prediction, with learning rate 1e-4.
Then, we use the full loss function to simultaneously update the parameters of $E_{\theta}$ and our morphable face reflectance model for 2 epochs, with learning rate 1e-5.
We adopt Adam optimizer~\cite{Kingma2015AdamAM}.

% \vspace{-5pt}
% \paragraph{Metrics}
% We report the SSIM~\cite{Wang2004ImageQA}, PSNR, and LPIPS~\cite{Zhang2018TheUE} scores in the face region to measure the image quality.
% We compute the metrics only in the face region.

\subsection{Evaluations}
% \vspace{-5pt}
\paragraph{Reflectance Parameter Estimation}
\begin{table}
    \centering
    \scriptsize
    \vspace{-5pt}
    \caption{Quantitative comparison of face rendering results on the Multi-PIE dataset between our method and previous works with a global specular exponent.}
    \begin{tabular}{lccc}
    \toprule
    & LPIPS~$\downarrow$ & SSIM~$\uparrow$ & PSNR~$\uparrow$ \\
    \midrule
    Exp=8 & 0.097 & 0.937 & 24.56  \\
    Exp=16 & 0.098 & 0.936 & 24.54  \\
    Exp=32 & 0.097 & 0.937 & 24.72  \\
    Exp=64 & 0.096 & 0.937 & 24.82  \\
    Ours & \textbf{0.083} & \textbf{0.942} & \textbf{25.00}  \\
    \bottomrule
    \end{tabular}
    \label{table:exp_cmp_global_shiness}
\end{table}

In our reflectance representation, we use the linear combination of 3 Blinn-Phong BRDFs with specular exponents 1, 8, and 64 as the specular term.
In Figure~\ref{Fig:exp_cmp_global_shiness}, we compare the face rendering results between our reflectance representation and previous works~\cite{dib2021practical,dib2021towards,dib2022s2f2} with spatially varying specular albedo and a global specular exponent. 
As discussed in Section~\ref{sec:method:representation}, we implement the reflectance representation of previous work by setting $k_{bp}\!\!=\!\!1$, and use the same inverse rendering pipeline described in Section~\ref{sec:method:init}.
The results illustrate that our representation obtains more realistic rendering results than the previous work. 
For example, the specular shiness is more significant at the tip of nose while vanishing around the cheek; previous works with global specular exponent cannot capture this phenomenon.
With the spatially varying linear combination weights of different Blinn-Phong BRDFs, our method can naturally represent spatially varying specular intensity and shiness.
In Table~\ref{table:exp_cmp_global_shiness}, we report the SSIM~\cite{Wang2004ImageQA}, PSNR, and LPIPS~\cite{Zhang2018TheUE} scores in the face region to quantitatively measure the discrepancy between the re-rendered face and the ground truth.
Again, our method achieves better results than the global exponent counterpart.

\vspace{-5pt}
\paragraph{Model Visualization}
We visualize the first 3 principal components of our morphable model in Figure~\ref{Fig:teaser}, including the diffuse albedo, and the weights for Blinn-Phong BRDF with specular exponents 1, 8, and 64, from left to right; we multiply the weights by 3 for better visualization.
It shows that our model learns to assign a large specular shiness to the tip of the nose while a small value to the cheek. 
See more visualizations in our \emph{Supplementary Material}.

\begin{table}
    \centering
    \scriptsize
    \vspace{-5pt}
    \caption{Photometric face reconstruction comparison between our method and competitors on 1000 images randomly sampled from the CelebA-HQ dataset.}
    \begin{tabular}{lcccccc}
    \toprule
    & LPIPS~$\downarrow$ & SSIM~$\uparrow$ & PSNR~$\uparrow$ \\
    \midrule
    BFM09 & 0.114 & 0.893 & 23.51 \\
    AlbedoMM & 0.116 & \textbf{0.901} & 23.69 \\
    Ours & \textbf{0.110} & 0.896 & \textbf{24.21}  \\
    \midrule
    Ours w/o finetune & 0.1256 & 0.886 & 23.26 \\
    \bottomrule
    \end{tabular}
    \label{table:exp_recon}
\end{table}

\begin{figure}[t]
    \centering
    \includegraphics[width=0.45\textwidth]{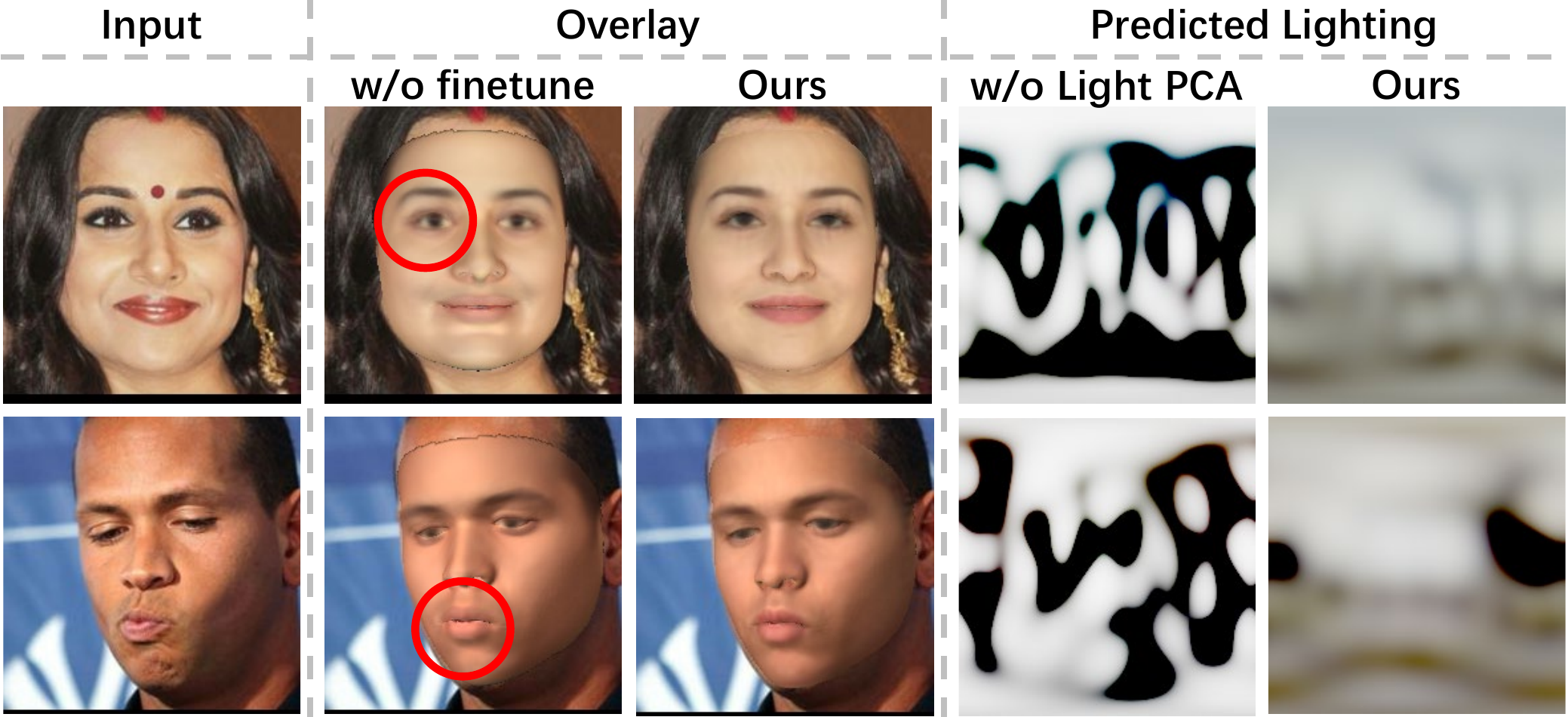}
    \vspace{-5pt}
    \caption{Qualitative ablation study of model finetuning and the use of our lighting PCA model.}
    \label{Fig:exp_ablat}
\end{figure}

\vspace{-5pt}
\paragraph{Ablation Study}
As shown in Figure~\ref{Fig:exp_ablat} and Table~\ref{table:exp_recon}, the proposed finetuning strategy can improve the generalization capability and expressive power of our initial morphable face reflectance model, leading to better face reconstruction quality, especially around mouth and eyes.
We then verify the effectiveness of our lighting PCA model. 
As illustrated in Figure~\ref{Fig:exp_ablat}, directly predicts 273 coefficients of the 8-order SH (w/o Light PCA) leads to unreasonable results; our lighting PCA model obtains better lighting predictions by constraining the searching space of the SH coefficients.

\subsection{Comparisons}
\paragraph{Baselines} We compare our method with BFM09~\cite{Paysan2009A3F} and AlbedoMM~\cite{smith2020morphable}.
BFM09 is a diffuse-only model built from 3D scans; AlbedoMM is a morphable model for diffuse and specular albedo built from Light Stage data.
To ensure a fair comparison, we use the same CNN-based framework (see Section~\ref{sec:method:finetune}) to implement the competitors. 
We train the reconstruction network for them on the FFHQ dataset, but we do not update their morphable model parameters.
As we only focus on appearance, the reconstruction network only predicts reflectance coefficients and lighting parameters and uses fixed precomputed geometry during training.
For BFM09, we adopt the same geometry parameters as ours.
For AlbedoMM, we use the same steps as mentioned in Section~\ref{sec:method:finetune} to obtain its geometry parameters.
Akin to~\cite{smith2020morphable}, we adopt the Blinn-Phong BRDF for AlbedoMM and set the global shiness to 20.

\begin{figure}[t]
    \centering
    \includegraphics[width=0.45\textwidth]{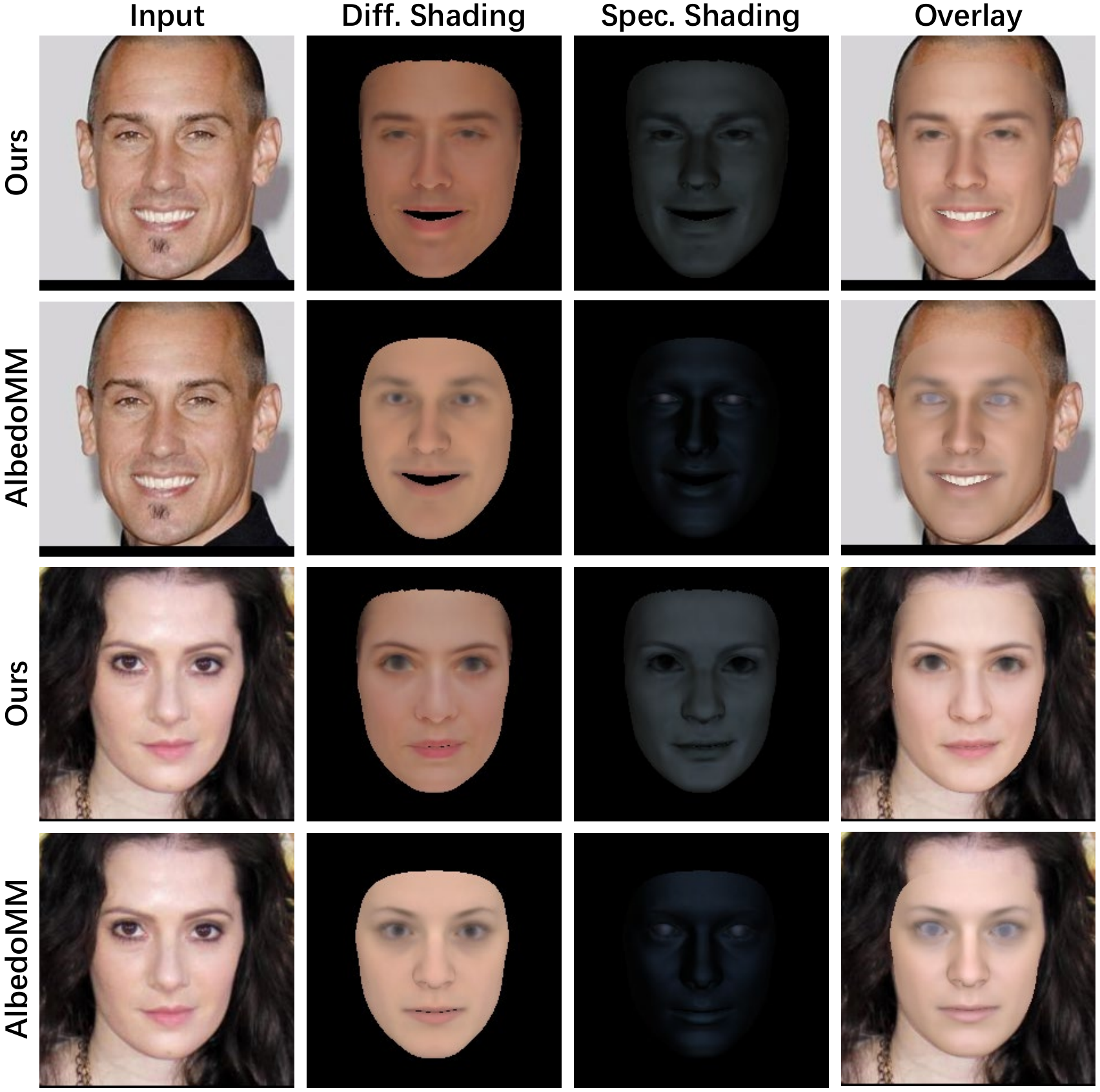}
    \vspace{-5pt}
    \caption{Qualitative comparison of face reconstruction and shading contributions between our method and AlbedoMM.}
    \label{Fig:exp_recon}
\end{figure}

\paragraph{Face Reconstruction}
We evaluate our method and the competitors on the CelebA-HQ~\cite{Karras2018ProgressiveGO} dataset.
% As illustrated in Figure~\ref{Fig:exp_recon} and Table~\ref{table:exp_recon}, our method can reconstruct more .  
As illustrated in Table~\ref{table:exp_recon}, our method obtains better photometric face reconstruction quantitative scores than the competitors since we finetune it on an in-the-wild dataset to improve its generalization capability and expressive power while the competitors are built from a limited number of scans.
As shown in Figure~\ref{Fig:exp_recon}, our method can reconstruct the input image well;
compared to AlbedoMM trained from Light Stage scans, our method trained from low-cost data can also disentangle the diffuse and specular shading in a plausible way.

\begin{figure}[t]
    \centering
    \includegraphics[width=0.45\textwidth]{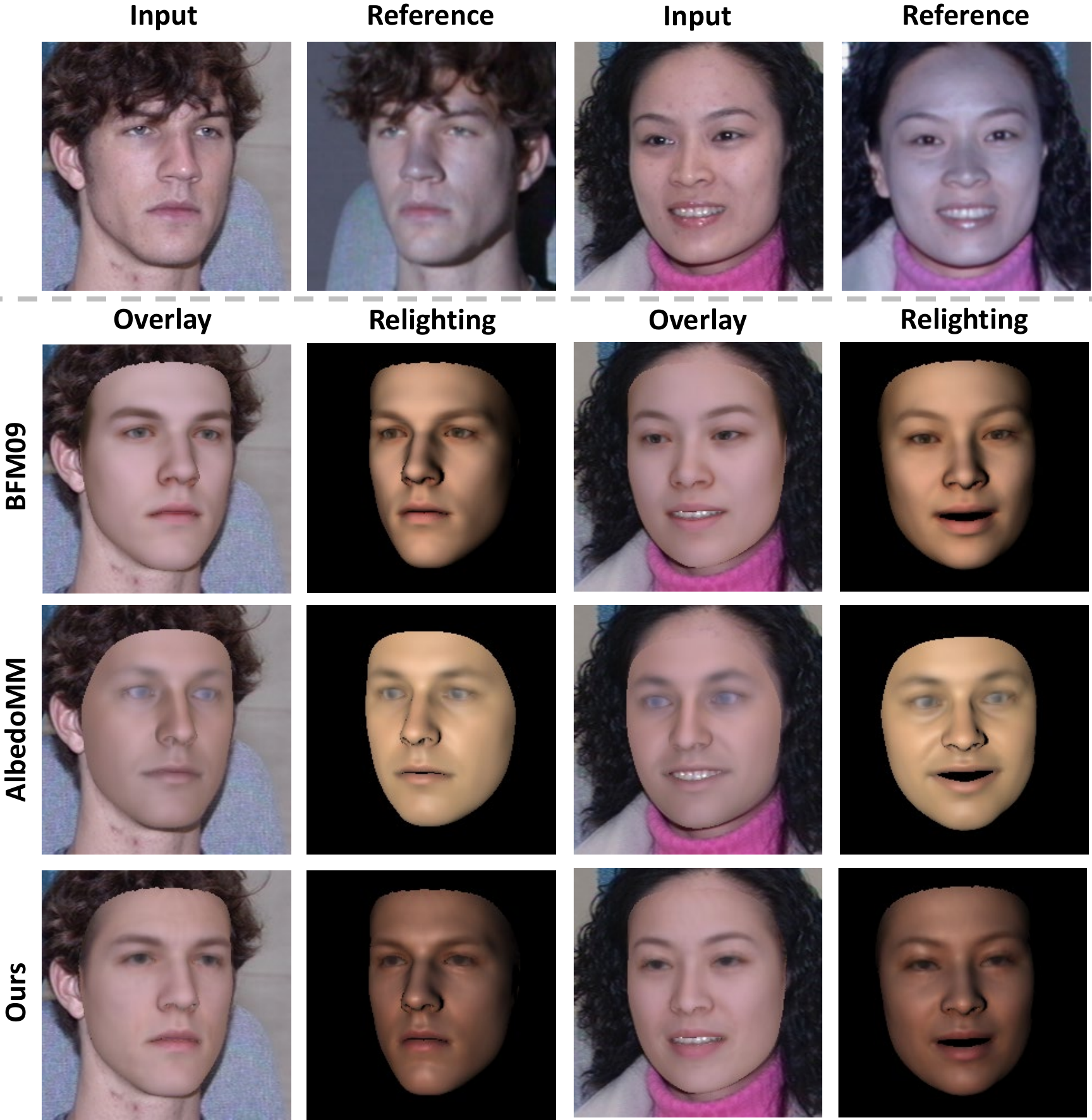}
    \vspace{-5pt}
    \caption{Qualitative comparison of face relighting results under point light source between our method and the competitors.}
    \label{Fig:exp_relight}
    \vspace{-5pt}
\end{figure}

\vspace{-5pt}
\paragraph{Face Relighting}
We evaluate the relighting performance of our method and the competitors on the Multi-PIE dataset~\cite{gross2010multi}.
Specifically, given an input image, we first obtain its geometry parameters as described in Section~\ref{sec:method:init} and reconstruct its reflectance parameters using our CNN-based face reconstruction network (columns 1 and 3 in Figure~\ref{Fig:exp_relight}).
Then, we re-render the image under a new point light source and compare it to the corresponding OLAT image with exactly the same light position obtained from the preprocessed Multi-PIE dataset in Section~\ref{sec:method:init} (columns 2 and 4 in Figure~\ref{Fig:exp_relight}). 
Since we do not have the ground truth light color and intensity information of the Multi-PIE dataset, we render our method and the competitors using the same white point light source for a fair comparison; \emph{please ignore the color difference and only focus on the distribution of facial specularities in Figure~\ref{Fig:exp_relight}}.
Compared to BFM09, our method successfully render plausible facial specularities since we adopt a non-Lambertian reflectance representation.
Compared to AlbedoMM, our method achieves more realistic results especially around the tip of the nose since we can model both spatially varying specular intensity and shiness.
See the video comparisons on our \href{https://yxuhan.github.io/ReflectanceMM/index.html}{\textcolor{magenta}{\emph{project page}}} for a better demonstration.

\begin{figure}[t]
    \centering
    \includegraphics[width=0.45\textwidth]{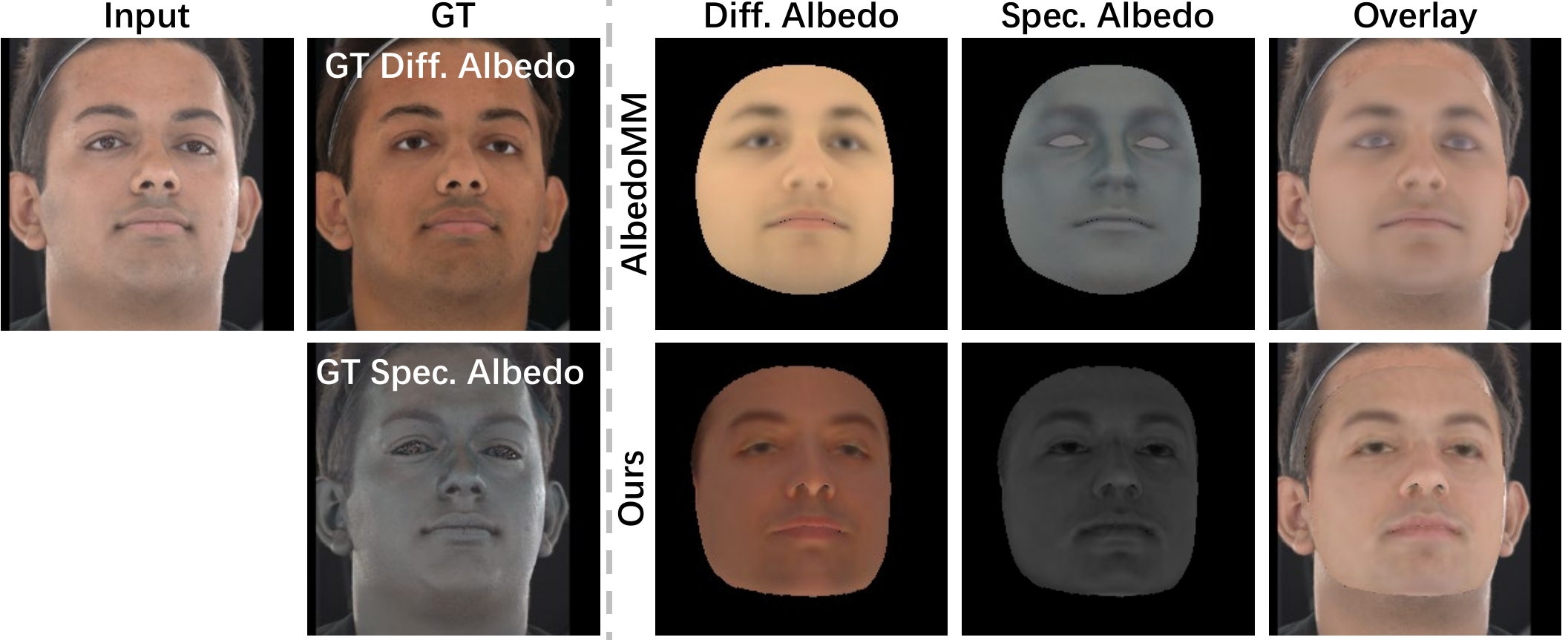}
    \vspace{-5pt}
    \caption{Qualitative comparison of diffuse and specular albedo reconstruction on the 3D-RFE dataset.}
    \label{Fig:exp_3drfe}
    \vspace{-5pt}
\end{figure}

\vspace{-5pt}
\paragraph{Reflectance Reconstruction}
Although our goal is not to model physically-accurate reflectance parameters, we compare our method with AlbedoMM on 23 Light Stage scans with ground truth diffuse and specular albedo captured under neutral expression from the 3D-RFE database~\cite{Stratou2011EffectOI}.
We adopt the sum of 3 linear combination weights in our reflectance representation as the specular albedo; this quantity shares the same meaning as the specular albedo, \emph{i.e.} the specular shading under a spatially-uniform environment lighting with unit radiance.
As shown in Figure~\ref{Fig:exp_3drfe}, our method can reconstruct plausible reflectance maps.
However, as shown in Table~\ref{table:exp_3drfe}, our method obtains inferior quantitative results than AlbedoMM on specular albedo reconstruction. 
We attribute this to two reasons: \emph{i)} AlbedoMM uses the 3D-RFE database to build their model~\cite{smith2020morphable} while our method has never seen these scans, and \emph{ii)} our method is built from low-cost data without lighting information, so there exists a global scale between our reflectance parameters and the ground truth although we try to mitigate it by setting a reasonable lighting color in Section~\ref{sec:method:init}.

\begin{table}
    \centering
    \scriptsize
    \vspace{-5pt}
    \caption{Quantitative diffuse and specular albedo reconstruction results on 23 Light Stage scans from the 3D-RFE dataset (PSNR$\uparrow$).}
    \begin{tabular}{lcccccc}
    \toprule
    & Diffuse Albedo & Specular Albedo\\
    \midrule
    AlbedoMM & 19.31 & \textbf{26.51} \\
    Ours & \textbf{20.13} & 19.14  \\
    \bottomrule
    \end{tabular}
    \label{table:exp_3drfe}
\end{table}

\section{Conclusion}
We propose the first 3D morphable face reflectance model with spatially varying BRDF, using only low-cost publicly-available data.
To represent spatially varying reflectance, we apply linear shiness weighting into parametric face modeling.
We develop an inverse rendering algorithm to reconstruct the reflectance parameters from the Multi-PIE dataset, from which we build an initial model.
We propose a strategy that jointly learns the face reconstruction network and updates the morphable model parameters on the FFHQ dataset to improve its generalization capability and expressive power.
Our method obtains decent rendering results with plausible facial specularities.
We discuss the limitations of our method in the \emph{Supplementary Material}.

\section*{Acknowledgement}
This work was supported by the National Key R\&D Program of China (2018YFA0704000), Beijing Natural Science Foundation (M22024), the NSFC (No.62021002), and the Key Research and Development Project of Tibet Autonomous Region (XZ202101ZY0019G). This work was also supported by THUIBCS, Tsinghua University, and BLBCI, Beijing Municipal Education Commission. We thank the anonymous reviewers for their constructive suggestions. Feng Xu is the corresponding author.

\appendix
\section{More Implementation Details}
\subsection{Multi-PIE Dataset Preprocessing}
We select 9 viewpoints (09\_0, 08\_0, 13\_0, 14\_0, 05\_1, 05\_0, 04\_1, 19\_0, and 20\_0) and 11 flashes (03, 04, 05, 06, 07, 08, 09, 10, 11, 14, and 18) for reflectance parameter estimation.
Please refer to~\cite{gross2010multi} for the detailed configuration of the viewpoints and flashes.
We develop a model-based method to reconstruct the camera parameters and the BFM09~\cite{Paysan2009A3F} geometry coefficients for each identity.
According to the Multi-PIE dataset~\cite{gross2010multi}, each selected viewpoint has one selected flash attached to it\footnote{We use the viewpoints 08\_1 and 19\_1 to solve the position of the flashes 14 and 18. However, we do not use the images captured by 08\_1 and 19\_1 since there exists apparent color inconsistency between these two viewpoints and the other selected 9 viewpoints.}. 
Hence, we approximate the flash position as the camera position.

We use the room-light images~\cite{gross2010multi} for reconstruction.
Specifically, we first adopt a CNN-based single-view face reconstruction method~\cite{deng2019accurate} to obtain the BFM09 coefficients, illumination coefficients, and head pose for each room-light image of a given identity. 
Then, we apply an offline optimization using the same loss function as~\cite{deng2019accurate} to improve the reconstruction accuracy.
During the offline optimization, each room-light image shares the same BFM09 coefficients since they are the multi-view images of the given identity, and we initialize them as the average of the coefficients of all views predicted by the face reconstruction CNN.
Similar to~\cite{deng2019accurate}, we use the perspective camera model with a reasonable predefined focal length to represent the 3D-2D projection.
After reconstruction, we can compute the camera parameters from the head pose $\textbf{R}$ and $\textbf{t}$ for each viewpoint:
\begin{equation}
    \textbf{R}_{cam} = \textbf{R}^{\rm T},\quad \textbf{t}_{cam} = -\textbf{R}^{\rm T}\cdot\textbf{t}
\end{equation}
Here, $\textbf{R}_{cam}$ and $\textbf{t}_{cam}$ are the camera rotation and translation in the BFM09 canonical space, respectively.
We repeat the steps above for all the identities in the Multi-PIE dataset.

Before reflectance parameter estimation, we obtain the OLAT image by removing the effect of the room light in the flash image. 
Specifically, we subtract the room-light image from the flash image in linear space with a reasonable mapping function\footnote{We empirically find that performing image differencing in linear space leads to better reflectance parameter estimation than in non-linear space.}:
\begin{equation}
    I_{OLAT} = (I_{flash})^{1.2} - (I_{roomlit})^{1.2}
\end{equation}
Here, $I_{OLAT}$ is the OLAT image in linear space, $I_{flash}$ and $I_{roomlit}$ are the flash and room-light image provided by the Multi-PIE dataset, respectively.
We then estimate the reflectance parameters from $I_{OLAT}$ and build our morphable face reflectance model in linear space.
To synthesis a face image in nonlinear space, we convert the shading $s$ to pixel color $c$ using the inverse mapping:
\begin{equation}
    c = s^{\frac{1}{1.2}}
\end{equation}

\vspace{-5pt}
\paragraph{Demographics}
Our initial morphable face reflectance model is built from a total of  128 manually selected individuals from the Multi-PIE dataset.
% We show the age distribution for the participants in Figure and skin type in Figure.
We release the ID of the selected individuals in our \href{https://github.com/yxuhan/ReflectanceMM}{\textcolor{magenta}{\emph{code repository}}}.

\vspace{-5pt}
\paragraph{Feasibility of reflectance parameter estimation}
The RGB diffuse color and 3 linear combination weights are the only unknowns in our reflectance representation. 
Theoretically, the ambiguity can be solved with 6 independent equations. 
We have 99 light-view direction pairs (the combination of 9 viewpoints and 11 light directions) in total, and if considering visibility,  most of the vertices have 50+ light-view direction pairs.
Different light-view direction pairs give independent equations.
Thus, it's feasible to estimate the BRDF parameters theoretically.

Practically, the light-view direction pairs which are not hitting the lobe of the BRDF would lead to a low activation value, and thus solving the reflectance parameters from these equations are highly ill-posed.
In our setup, we find that the ill-posed scenario only happens on very few face vertices on the side face or with normal directions going down like nares.
For most of the face vertices, our setup can provide enough well-conditioned equations with the corresponding light-view direction pairs hitting the lobe.
Thus, it's feasible to estimate the BRDF parameters practically.

\subsection{Model Finetuning}
Recall that in model finetuning, the learnable parameters are the morphable model parameters, including the mean $\bar{R}$ and bases ${\rm M_R}$, and face reconstruction network parameters $\theta$. 
We optimize them with the combination of a reconstruction loss $\mathcal{L}_{rec}$ and a regularization loss $\mathcal{L}_{reg}$:
\begin{equation}
    \mathop{\arg\min}\limits_{\bar{R},{\rm M_R},\theta} \mathcal{L}_{rec} + \mathcal{L}_{reg}
\end{equation}

$\mathcal{L}_{rec}$ is the combination of a L1 term $\mathcal{L}_{l1}$ and a perceptual term $\mathcal{L}_{per}$:
\begin{align}
    \mathcal{L}_{rec} &= \omega_{l1}\cdot\mathcal{L}_{l1} + \omega_{per}\cdot\mathcal{L}_{per}, {\rm where} \\
    \mathcal{L}_{l1} &= M_{skin} \cdot ||\hat{I}-I||_1\\
    \mathcal{L}_{per} &= 1 - \langle \phi_{feat}(\hat{I}), \phi_{feat}(I) \rangle
\end{align}
Here, $M_{skin}$ is the mask indicated skin region, obtained by an off-the-shelf face parsing method~\cite{yu2021bisenet}; $\langle\cdot,\cdot\rangle$ is the inner product operation; $\phi_{feat}$ is a pretrained FaceNet architecture~\cite{Schroff2015FaceNetAU} for feature extraction.
Note that we directly compute the reconstruction loss $\mathcal{L}_{rec}$ in the linear space.
Although $\phi_{feat}$ is trained on images in the nonlinear space, we empirically find that it can still provide a reasonable supervision signal if the input image is in the linear space.

In our regularization loss $\mathcal{L}_{reg}$, we first adopt $\mathcal{L}_{coef}$ to constrain the predicted PCA coefficients $\beta$ and $\gamma$:
\begin{equation}
    \mathcal{L}_{coef} = \sum_{i=1}^{N_R} (\frac{\beta_i}{\sigma_{\beta_i}})^2 + \sum_{i=1}^{N_L} (\frac{\gamma_i}{\sigma_{\gamma_i}})^2
\end{equation}
Here, $\sigma_{\beta}$ and $\sigma_{\gamma}$ are the standard deviations of the initial morphable face reflectance model and the lighting PCA model, respectively.
 Then, to constrain the updating of our morphable reflectance model, we design $\mathcal{L}_{upd}$ as:
\begin{equation}
    \mathcal{L}_{upd} = ||\bar{R}-\bar{R}_0||_1 + ||\rm{M_R} - \rm{M_{R_0}}||_1 
\end{equation}
Here, $\bar{R}_0$ and $\rm{M_{R_0}}$ are the mean and bases of our initial morphable face reflectance model built from the Multi-PIE dataset.
To resolve the color ambiguity between albedo and lighting, we involve $\mathcal{L}_{light}$ to encourage monochromatic environment lighting as~\cite{DFRgithub}:
\begin{equation}
    \mathcal{L}_{light} = ||l-l_{mean}||_2^2
\end{equation}
Here, $l$ is the retrieved 8-$th$ order SH coefficients; $l_{mean}$ is the mean of $l$ over the color channel dimension, representing the monochromatic counterpart of $l$. 
Thus, our regularization loss $\mathcal{L}_{reg}$ can be written as:
\begin{equation}
    \mathcal{L}_{reg} = \omega_{coef}\cdot\mathcal{L}_{coef} + \omega_{upd}\cdot\mathcal{L}_{upd} + \omega_{light}\cdot\mathcal{L}_{light}
\end{equation}

In our experiments, we set $\omega_{l1}$, $\omega_{per}$, $\omega_{coef}$, $\omega_{upd}$, $\omega_{light}$ to 2, 0.1, 0.001, 10, 10, respectively.

\section{More Results}
\subsection{Model Visualization}
\begin{figure*}[t]
    \centering
    \includegraphics[width=1\textwidth]{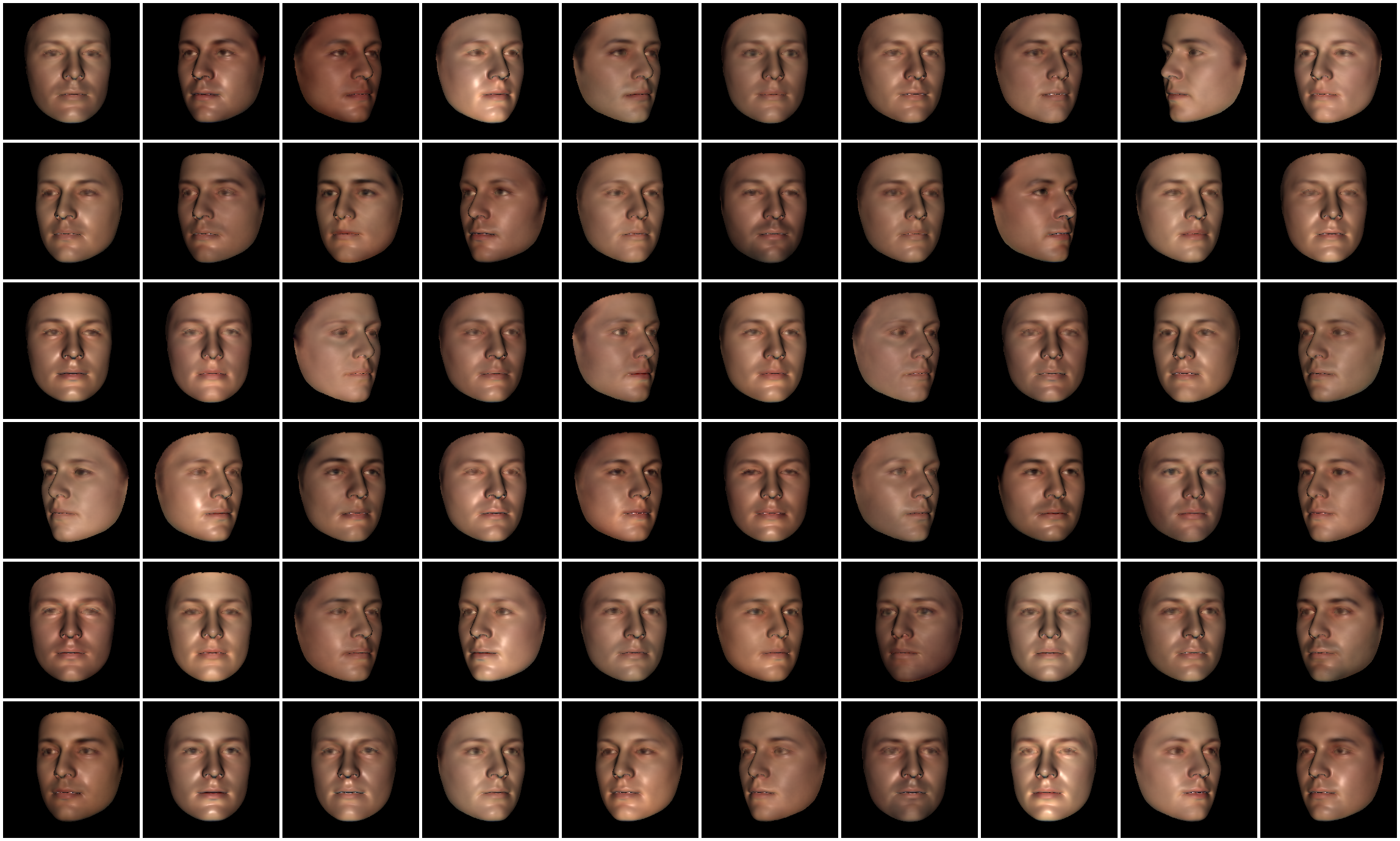}
    \vspace{-5pt}
    \caption{60 random samples drawn from our initial morphable face reflectance model (\textbf{\emph{before model finetuning}}). Rendered in nonlinear sRGB space with a white frontal point light.}
    \label{Fig:rand_sample_before}
\end{figure*}

\begin{figure*}[t]
    \centering
    \includegraphics[width=1\textwidth]{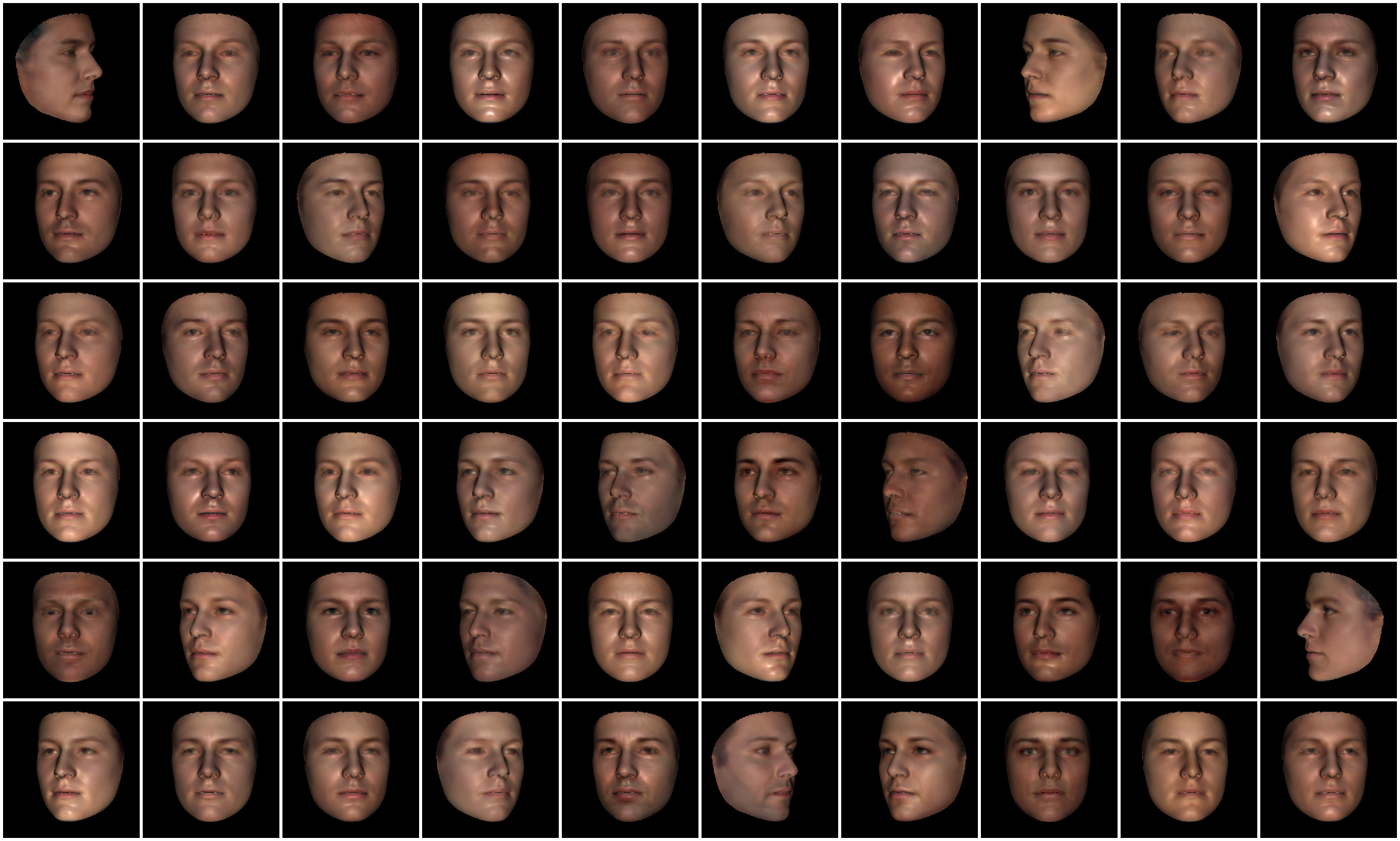}
    \vspace{-5pt}
    \caption{60 random samples drawn from our final morphable face reflectance model (\textbf{\emph{after model finetuning}}). Rendered in nonlinear sRGB space with a white frontal point light.}
    \label{Fig:rand_sample_after}
\end{figure*}

In Figure~\ref{Fig:rand_sample_before} and Figure~\ref{Fig:rand_sample_after}, we visualize our model by showing random samples drawn from it before and after fine-tuning, respectively.
The images are rendered in nonlinear space with a white frontal point light.

\begin{figure*}[t]
    \centering
    \includegraphics[width=1\textwidth]{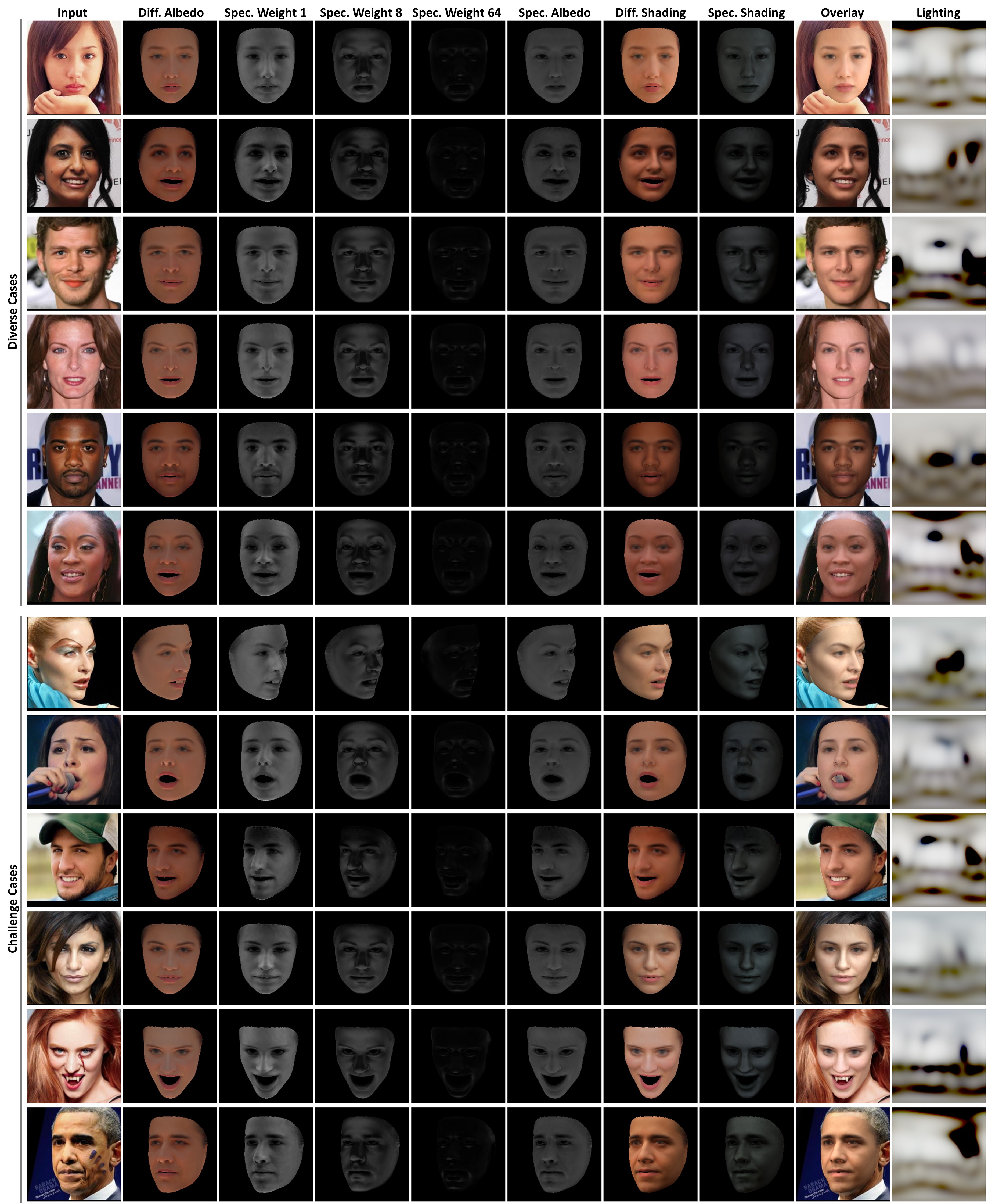}
    \vspace{-5pt}
    \caption{Face reconstruction results on diverse in-the-wild face images.}
    \label{Fig:in-the-wild}
\end{figure*}

\subsection{Face Reconstruction}

\vspace{-5pt}
\paragraph{More Reconstruction Results}
We show more face reconstruction results on in-the-wild face images in Figure~\ref{Fig:in-the-wild}, including diverse ethics groups and challenging cases with facial occlusions and makeups. 
We multiply the linear combination weights (columns 3, 4 5 in Figure~\ref{Fig:in-the-wild}) by 3  for better visualization.

Thanks to the model-finetuning process, our method is robust to handle diverse input images and predicts plausible reflectance attributes.
However, it has the same limitation as previous in-the-wild face reconstruction methods~\cite{deng2019accurate,tewari2021learning,tewari2019fml}: \emph{i)} the global skin tone can not be disentangled from the illumination due to the scale ambiguity between lighting and reflectance (row 5), and \emph{ii)} shadow cast by external geometry (hat in row 9) bakes into the reflectance channels.

\vspace{-5pt}
\paragraph{Evaluation on Geometry Reconstruction}

\begin{table}
    \centering
    \scriptsize
    \vspace{-5pt}
    \caption{Quantative face geometry reconstruction error on the validation set of the NoW challenge.}
    \begin{tabular}{lcccccc}
    \toprule
    & Median (mm)~$\downarrow$ & mean (mm)~$\downarrow$ & std (mm)~$\downarrow$ \\
    \midrule
    BFM09 & 1.44 & 2.06 & 2.51 \\
    Ours & 1.51 & 2.15 & 2.61 \\
    \bottomrule
    \end{tabular}
    \label{table:exp_geo_recon}
\end{table}

Although our goal is not to better reconstruct face shape from images, we compare our method and BFM09~\cite{Paysan2009A3F} on the validation set of the NoW challenge~\cite{RingNet:CVPR:2019} to help the readers better understand our model.
Note that both methods use the same BFM09 geometry model; we do not compare to AlbedoMM since AlbedoMM~\cite{smith2020morphable} is built on top of the BFM17~\cite{gerig2018morphable} geometry model.

In this experiment, we adopt a similar network architecture as~\cite{deng2019accurate} by simply modifying the number of neurons of the last fully-connect layer of $E_\theta(\cdot)$ from $N_R+N_L+3$ to $N_S+N_E+N_P+N_R+N_L+3$ to predict the shape and expression coefficients and the head pose.
We use the first 80 and 64 bases of the BFM09 shape and expression morphable model, respectively; thus, $N_S=80$ and $N_E=64$.
For the head pose, we use the Euler angle to represent rotation and a 3D vector to represent translation; thus, $N_P=6$.
To train the network for geometry reconstruction, we involve a landmark loss term akin to previous works~\cite{deng2019accurate,tewari2017mofa,tewari2018self,RingNet:CVPR:2019}:
\begin{equation}
    \mathcal{L}_{ldm} = \sum_{n=1}^{68} ||\hat{q_n}-q_n||_2^2
\end{equation}
Here, $q_n$ are the 2D landmarks obtained from an off-the-shelf landmark detector~\cite{bulat2017far}; $\hat{q_n}$ are the 2D projection of the 3D landmarks defined on the reconstructed shape.
In addition, we modify $\mathcal{L}_{coef}$ to add constraints on the shape and expression coefficients:
\begin{equation}
    \mathcal{L}_{coef} = \sum_{i=1}^{N_S} (\frac{\alpha_i}{\sigma_{\alpha_i}})^2 + \sum_{i=1}^{N_E} (\frac{\delta_i}{\sigma_{\delta_i}})^2 + \sum_{i=1}^{N_R} (\frac{\beta_i}{\sigma_{\beta_i}})^2 + \sum_{i=1}^{N_L} (\frac{\gamma_i}{\sigma_{\gamma_i}})^2
\end{equation}
Here, $\alpha\in\mathbb{R}^{N_S}$ and $\delta\in\mathbb{R}^{N_E}$ are the predicted shape and expression coefficients, respectively; $\sigma_{\alpha}$ and $\sigma_{\delta}$ are the standard deviations of the shape and expression morphable model, respectively.
Our full loss functions for geometry reconstruction can be written as:
\begin{align}
    \mathcal{L} &= \omega_{l1}\cdot\mathcal{L}_{l1} + \omega_{per}\cdot\mathcal{L}_{per}  \nonumber \\
    &+ \omega_{coef}\cdot\mathcal{L}_{coef} + \omega_{light}\cdot\mathcal{L}_{light} + \omega_{ldm}\cdot\mathcal{L}_{ldm}
\end{align}
In the geometry reconstruction experiments, we set $\omega_{l1}$, $\omega_{per}$, $\omega_{coef}$, $\omega_{light}$, $\omega_{ldm}$ to 2, 0.2, 0.001, 10, 0.002, respectively.
We train the geometry reconstruction network on the FFHQ~\cite{Karras2018ASG} dataset for 20 epochs.

As shown in Table~\ref{table:exp_geo_recon}, our method just obtains similar quantitative results compared to the BFM09 under the same CNN-based face geometry reconstruction pipeline.
However, we believe that our model has the potential to achieve better geometry reconstruction results with the advance of lighting estimation and differentiable ray tracer.

\subsection{Face Relighting and OLAT Rendering}
See our \href{https://yxuhan.github.io/ReflectanceMM/index.html}{\textcolor{magenta}{\emph{project page}}} for the video results.

\section{Limitations and Discussions}
Our method still has several limitations.
We adopt the Lambertian BRDF to represent diffuse reflectance.
Thus, we cannot model the subsurface scattering effect.
Integrating a more complicated reflectance representation~\cite{weyrich2006analysis} into our morphable face reflectance model to improve face rendering realism is an interesting direction.

Our model cannot well represent the specularities around the eyes.
We try a straightforward way by adding more mirror-like specular terms in our reflectance representation but find it does not work.
We attribute this to the following two reasons: \emph{i)} the reconstructed geometry is inaccurate around eyes during inverse rendering, and \emph{ii)} our BRDF reflectance representation cannot well model the complex properties of eyes (e.g. refraction).

During model finetuning, we use a differentiable rasterizer with an efficient local shading technique to render the reconstructed image, without considering global illumination effects like self-shadowing, considering that the illumination is soft, and the self-shadows are insignificant in most in-the-wild images.
We believe that using a differentiable ray tracer~\cite{Li:2018:DMC} would slightly improve the current results as demonstrated in existing works~\cite{dib2022s2f2,dib2021practical,dib2021towards}.
Moreover, leveraging a multi-view in-the-wild face image dataset~\cite{cao2018vggface2} or video dataset~\cite{chung2018voxceleb2} could improve the face reconstruction results, as demonstrated by the previous works~\cite{feng2021learning,tewari2019fml}.
We leave these as our future works.

In addition, there is an inevitably global scale between the reflectance parameters in our model and the ground truth since the low-cost data does not provide lighting information~\cite{ramamoorthi2001signal}.

%%%%%%%%% REFERENCES
{\small
\bibliographystyle{ieee_fullname}
\bibliography{PaperForReview}

\begin{thebibliography}{10}\itemsep=-1pt

\bibitem{akenine2019real}
Tomas Akenine-Moller, Eric Haines, and Naty Hoffman.
\newblock {\em Real-time rendering}.
\newblock AK Peters/crc Press, 2019.

\bibitem{Azinovic2022HighResFA}
Dejan Azinovi'c, Olivier Maury, Christophe Hery, Mathias Niessner, and Justus
  Thies.
\newblock High-res facial appearance capture from polarized smartphone images.
\newblock {\em ArXiv}, abs/2212.01160, 2022.

\bibitem{bai2021riggable}
Ziqian Bai, Zhaopeng Cui, Xiaoming Liu, and Ping Tan.
\newblock Riggable 3d face reconstruction via in-network optimization.
\newblock In {\em Proceedings of the IEEE/CVF Conference on Computer Vision and
  Pattern Recognition}, pages 6216--6225, 2021.

\bibitem{blanz1999morphable}
Volker Blanz and Thomas Vetter.
\newblock A morphable model for the synthesis of 3d faces.
\newblock In {\em Proceedings of the 26th annual conference on Computer
  graphics and interactive techniques}, pages 187--194, 1999.

\bibitem{blinn1977models}
James~F Blinn.
\newblock Models of light reflection for computer synthesized pictures.
\newblock In {\em Proceedings of the 4th annual conference on Computer graphics
  and interactive techniques}, pages 192--198, 1977.

\bibitem{Booth2016A3M}
James Booth, Anastasios Roussos, Stefanos Zafeiriou, Allan Ponniah, and
  David~J. Dunaway.
\newblock A 3d morphable model learnt from 10,000 faces.
\newblock {\em 2016 IEEE Conference on Computer Vision and Pattern Recognition
  (CVPR)}, pages 5543--5552, 2016.

\bibitem{bulat2017far}
Adrian Bulat and Georgios Tzimiropoulos.
\newblock How far are we from solving the 2d \& 3d face alignment problem? (and
  a dataset of 230,000 3d facial landmarks).
\newblock In {\em International Conference on Computer Vision}, 2017.

\bibitem{cao20133d}
Chen Cao, Yanlin Weng, Stephen Lin, and Kun Zhou.
\newblock 3d shape regression for real-time facial animation.
\newblock {\em ACM Transactions on Graphics (TOG)}, 32(4):1--10, 2013.

\bibitem{Cao2014FaceWarehouseA3}
Chen Cao, Yanlin Weng, Shun Zhou, Y. Tong, and Kun Zhou.
\newblock Facewarehouse: A 3d facial expression database for visual computing.
\newblock {\em IEEE Transactions on Visualization and Computer Graphics},
  20:413--425, 2014.

\bibitem{cao2018vggface2}
Qiong Cao, Li Shen, Weidi Xie, Omkar~M Parkhi, and Andrew Zisserman.
\newblock Vggface2: A dataset for recognising faces across pose and age.
\newblock In {\em 2018 13th IEEE international conference on automatic face \&
  gesture recognition (FG 2018)}, pages 67--74. IEEE, 2018.

\bibitem{chaudhuri2020personalized}
Bindita Chaudhuri, Noranart Vesdapunt, Linda Shapiro, and Baoyuan Wang.
\newblock Personalized face modeling for improved face reconstruction and
  motion retargeting.
\newblock In {\em European Conference on Computer Vision}, pages 142--160.
  Springer, 2020.

\bibitem{chung2018voxceleb2}
Joon~Son Chung, Arsha Nagrani, and Andrew Zisserman.
\newblock Voxceleb2: Deep speaker recognition.
\newblock {\em arXiv preprint arXiv:1806.05622}, 2018.

\bibitem{Dai2017A3M}
Hang Dai, Nick~E. Pears, William Smith, and Christian Duncan.
\newblock A 3d morphable model of craniofacial shape and texture variation.
\newblock {\em 2017 IEEE International Conference on Computer Vision (ICCV)},
  pages 3104--3112, 2017.

\bibitem{debevec2000acquiring}
Paul Debevec, Tim Hawkins, Chris Tchou, Haarm-Pieter Duiker, Westley Sarokin,
  and Mark Sagar.
\newblock Acquiring the reflectance field of a human face.
\newblock In {\em Proceedings of the 27th annual conference on Computer
  graphics and interactive techniques}, pages 145--156, 2000.

\bibitem{DFRgithub}
Yu Deng.
\newblock Deep3dfacereconstruction, 2019.
\newblock https://github.com/microsoft/Deep3DFaceReconstruction.

\bibitem{deng2020disentangled}
Yu Deng, Jiaolong Yang, Dong Chen, Fang Wen, and Xin Tong.
\newblock Disentangled and controllable face image generation via 3d
  imitative-contrastive learning.
\newblock In {\em Proceedings of the IEEE/CVF conference on computer vision and
  pattern recognition}, pages 5154--5163, 2020.

\bibitem{deng2019accurate}
Yu Deng, Jiaolong Yang, Sicheng Xu, Dong Chen, Yunde Jia, and Xin Tong.
\newblock Accurate 3d face reconstruction with weakly-supervised learning: From
  single image to image set.
\newblock In {\em Proceedings of the IEEE/CVF Conference on Computer Vision and
  Pattern Recognition Workshops}, pages 0--0, 2019.

\bibitem{dib2022s2f2}
Abdallah Dib, Junghyun Ahn, Cedric Thebault, Philippe-Henri Gosselin, and Louis
  Chevallier.
\newblock S2f2: Self-supervised high fidelity face reconstruction from
  monocular image.
\newblock {\em arXiv preprint arXiv:2203.07732}, 2022.

\bibitem{dib2021practical}
Abdallah Dib, Gaurav Bharaj, Junghyun Ahn, C{\'e}dric Th{\'e}bault, Philippe
  Gosselin, Marco Romeo, and Louis Chevallier.
\newblock Practical face reconstruction via differentiable ray tracing.
\newblock In {\em Computer Graphics Forum}, volume~40, pages 153--164. Wiley
  Online Library, 2021.

\bibitem{dib2021towards}
Abdallah Dib, C{\'e}dric Th{\'e}bault, Junghyun Ahn, Philippe-Henri Gosselin,
  Christian Theobalt, and Louis Chevallier.
\newblock Towards high fidelity monocular face reconstruction with rich
  reflectance using self-supervised learning and ray tracing.
\newblock In {\em Proceedings of the IEEE/CVF International Conference on
  Computer Vision}, pages 12819--12829, 2021.

\bibitem{egger2018occlusion}
Bernhard Egger, Sandro Sch{\"o}nborn, Andreas Schneider, Adam Kortylewski,
  Andreas Morel-Forster, Clemens Blumer, and Thomas Vetter.
\newblock Occlusion-aware 3d morphable models and an illumination prior for
  face image analysis.
\newblock {\em International Journal of Computer Vision}, 126(12):1269--1287,
  2018.

\bibitem{egger20203d}
Bernhard Egger, William~AP Smith, Ayush Tewari, Stefanie Wuhrer, Michael
  Zollhoefer, Thabo Beeler, Florian Bernard, Timo Bolkart, Adam Kortylewski,
  Sami Romdhani, et~al.
\newblock 3d morphable face models—past, present, and future.
\newblock {\em ACM Transactions on Graphics (TOG)}, 39(5):1--38, 2020.

\bibitem{feng2021learning}
Yao Feng, Haiwen Feng, Michael~J Black, and Timo Bolkart.
\newblock Learning an animatable detailed 3d face model from in-the-wild
  images.
\newblock {\em ACM Transactions on Graphics (ToG)}, 40(4):1--13, 2021.

\bibitem{garrido2016reconstruction}
Pablo Garrido, Michael Zollh{\"o}fer, Dan Casas, Levi Valgaerts, Kiran
  Varanasi, Patrick P{\'e}rez, and Christian Theobalt.
\newblock Reconstruction of personalized 3d face rigs from monocular video.
\newblock {\em ACM Transactions on Graphics (TOG)}, 35(3):1--15, 2016.

\bibitem{genova2018unsupervised}
Kyle Genova, Forrester Cole, Aaron Maschinot, Aaron Sarna, Daniel Vlasic, and
  William~T Freeman.
\newblock Unsupervised training for 3d morphable model regression.
\newblock In {\em Proceedings of the IEEE Conference on Computer Vision and
  Pattern Recognition}, pages 8377--8386, 2018.

\bibitem{gerig2018morphable}
Thomas Gerig, Andreas Morel-Forster, Clemens Blumer, Bernhard Egger, Marcel
  Luthi, Sandro Sch{\"o}nborn, and Thomas Vetter.
\newblock Morphable face models-an open framework.
\newblock In {\em 2018 13th IEEE International Conference on Automatic Face \&
  Gesture Recognition (FG 2018)}, pages 75--82. IEEE, 2018.

\bibitem{ghosh2011multiview}
Abhijeet Ghosh, Graham Fyffe, Borom Tunwattanapong, Jay Busch, Xueming Yu, and
  Paul Debevec.
\newblock Multiview face capture using polarized spherical gradient
  illumination.
\newblock In {\em Proceedings of the 2011 SIGGRAPH Asia Conference}, pages
  1--10, 2011.

\bibitem{gross2010multi}
Ralph Gross, Iain Matthews, Jeffrey Cohn, Takeo Kanade, and Simon Baker.
\newblock Multi-pie.
\newblock {\em Image and vision computing}, 28(5):807--813, 2010.

\bibitem{He2016DeepRL}
Kaiming He, X. Zhang, Shaoqing Ren, and Jian Sun.
\newblock Deep residual learning for image recognition.
\newblock {\em 2016 IEEE Conference on Computer Vision and Pattern Recognition
  (CVPR)}, pages 770--778, 2016.

\bibitem{kajiya1986rendering}
James~T Kajiya.
\newblock The rendering equation.
\newblock In {\em Proceedings of the 13th annual conference on Computer
  graphics and interactive techniques}, pages 143--150, 1986.

\bibitem{Karras2018ProgressiveGO}
Tero Karras, Timo Aila, Samuli Laine, and Jaakko Lehtinen.
\newblock Progressive growing of gans for improved quality, stability, and
  variation.
\newblock {\em ArXiv}, abs/1710.10196, 2018.

\bibitem{Karras2018ASG}
Tero Karras, Samuli Laine, and Timo Aila.
\newblock A style-based generator architecture for generative adversarial
  networks.
\newblock {\em 2019 IEEE/CVF Conference on Computer Vision and Pattern
  Recognition (CVPR)}, pages 4396--4405, 2018.

\bibitem{karras2020analyzing}
Tero Karras, Samuli Laine, Miika Aittala, Janne Hellsten, Jaakko Lehtinen, and
  Timo Aila.
\newblock Analyzing and improving the image quality of stylegan.
\newblock In {\em Proceedings of the IEEE/CVF conference on computer vision and
  pattern recognition}, pages 8110--8119, 2020.

\bibitem{Kingma2015AdamAM}
Diederik~P. Kingma and Jimmy Ba.
\newblock Adam: A method for stochastic optimization.
\newblock {\em CoRR}, abs/1412.6980, 2015.

\bibitem{Klehm2015RecentAI}
Oliver Klehm, Fabrice Rousselle, Marios Papas, Derek Bradley, Christophe Hery,
  B. Bickel, Wojciech Jarosz, and Thabo Beeler.
\newblock Recent advances in facial appearance capture.
\newblock {\em Computer Graphics Forum}, 34, 2015.

\bibitem{Kumar2010MorphableRF}
Ritwik~K. Kumar, Michael~J. Jones, and Tim~K. Marks.
\newblock Morphable reflectance fields for enhancing face recognition.
\newblock {\em 2010 IEEE Computer Society Conference on Computer Vision and
  Pattern Recognition}, pages 2606--2613, 2010.

\bibitem{Laine2020ModularPF}
Samuli Laine, Janne Hellsten, Tero Karras, Yeongho Seol, Jaakko Lehtinen, and
  Timo Aila.
\newblock Modular primitives for high-performance differentiable rendering.
\newblock {\em ACM Transactions on Graphics (TOG)}, 39:1 -- 14, 2020.

\bibitem{Lattas2020AvatarMeRR}
Alexandros Lattas, Stylianos Moschoglou, Baris Gecer, Stylianos Ploumpis,
  Vasileios Triantafyllou, Abhijeet Ghosh, and Stefanos Zafeiriou.
\newblock Avatarme: Realistically renderable 3d facial reconstruction
  “in-the-wild”.
\newblock {\em 2020 IEEE/CVF Conference on Computer Vision and Pattern
  Recognition (CVPR)}, pages 757--766, 2020.

\bibitem{Li2014IntrinsicFI}
Chen Li, Kun Zhou, and Stephen Lin.
\newblock Intrinsic face image decomposition with human face priors.
\newblock In {\em European Conference on Computer Vision}, 2014.

\bibitem{Li2020LearningFO}
Ruilong Li, Kalle Bladin, Yajie Zhao, Chinmay Chinara, Owen Ingraham, Pengda
  Xiang, Xinglei Ren, Pratusha~Bhuvana Prasad, Bipin Kishore, Jun Xing, and Hao
  Li.
\newblock Learning formation of physically-based face attributes.
\newblock {\em 2020 IEEE/CVF Conference on Computer Vision and Pattern
  Recognition (CVPR)}, pages 3407--3416, 2020.

\bibitem{li2017learning}
Tianye Li, Timo Bolkart, Michael~J Black, Hao Li, and Javier Romero.
\newblock Learning a model of facial shape and expression from 4d scans.
\newblock {\em ACM Trans. Graph.}, 36(6):194--1, 2017.

\bibitem{Li:2018:DMC}
Tzu-Mao Li, Miika Aittala, Fr{\'e}do Durand, and Jaakko Lehtinen.
\newblock Differentiable monte carlo ray tracing through edge sampling.
\newblock {\em ACM Trans. Graph. (Proc. SIGGRAPH Asia)}, 37(6):222:1--222:11,
  2018.

\bibitem{ma2007rapid}
Wan-Chun Ma, Tim Hawkins, Pieter Peers, Charles-Felix Chabert, Malte Weiss,
  Paul~E Debevec, et~al.
\newblock Rapid acquisition of specular and diffuse normal maps from polarized
  spherical gradient illumination.
\newblock {\em Rendering Techniques}, 2007(9):10, 2007.

\bibitem{MallikarjunB2021MonocularRO}
R. MallikarjunB., Ayush Tewari, Tae-Hyun Oh, Tim Weyrich, B. Bickel, Hans-Peter
  Seidel, Hanspeter Pfister, Wojciech Matusik, Mohamed~A. Elgharib, and
  Christian Theobalt.
\newblock Monocular reconstruction of neural face reflectance fields.
\newblock {\em 2021 IEEE/CVF Conference on Computer Vision and Pattern
  Recognition (CVPR)}, pages 4789--4798, 2021.

\bibitem{matusik2003data}
Wojciech Matusik.
\newblock {\em A data-driven reflectance model}.
\newblock PhD thesis, Massachusetts Institute of Technology, 2003.

\bibitem{pandey2021total}
Rohit Pandey, Sergio~Orts Escolano, Chloe Legendre, Christian Haene, Sofien
  Bouaziz, Christoph Rhemann, Paul Debevec, and Sean Fanello.
\newblock Total relighting: learning to relight portraits for background
  replacement.
\newblock {\em ACM Transactions on Graphics (TOG)}, 40(4):1--21, 2021.

\bibitem{Paysan2009A3F}
Pascal Paysan, Reinhard Knothe, Brian Amberg, Sami Romdhani, and Thomas Vetter.
\newblock A 3d face model for pose and illumination invariant face recognition.
\newblock {\em 2009 Sixth IEEE International Conference on Advanced Video and
  Signal Based Surveillance}, pages 296--301, 2009.

\bibitem{ramamoorthi2001efficient}
Ravi Ramamoorthi and Pat Hanrahan.
\newblock An efficient representation for irradiance environment maps.
\newblock In {\em Proceedings of the 28th annual conference on Computer
  graphics and interactive techniques}, pages 497--500, 2001.

\bibitem{ramamoorthi2001signal}
Ravi Ramamoorthi and Pat Hanrahan.
\newblock A signal-processing framework for inverse rendering.
\newblock In {\em Proceedings of the 28th annual conference on Computer
  graphics and interactive techniques}, pages 117--128, 2001.

\bibitem{Ravi2020Accelerating3D}
Nikhila Ravi, Jeremy Reizenstein, David Novotn{\'y}, Taylor Gordon, Wan-Yen Lo,
  Justin Johnson, and Georgia Gkioxari.
\newblock Accelerating 3d deep learning with pytorch3d.
\newblock {\em SIGGRAPH Asia 2020 Courses}, 2020.

\bibitem{Riviere2020SingleshotHF}
J{\'e}r{\'e}my Riviere, Paulo F.~U. Gotardo, Derek Bradley, Abhijeet Ghosh, and
  Thabo Beeler.
\newblock Single-shot high-quality facial geometry and skin appearance capture.
\newblock {\em ACM Transactions on Graphics (TOG)}, 39:81:1 -- 81:12, 2020.

\bibitem{RingNet:CVPR:2019}
Soubhik Sanyal, Timo Bolkart, Haiwen Feng, and Michael Black.
\newblock Learning to regress {3D} face shape and expression from an image
  without {3D} supervision.
\newblock In {\em Proceedings IEEE Conf. on Computer Vision and Pattern
  Recognition (CVPR)}, pages 7763--7772, June 2019.

\bibitem{Schroff2015FaceNetAU}
Florian Schroff, Dmitry Kalenichenko, and James Philbin.
\newblock Facenet: A unified embedding for face recognition and clustering.
\newblock {\em 2015 IEEE Conference on Computer Vision and Pattern Recognition
  (CVPR)}, pages 815--823, 2015.

\bibitem{Sengupta2021ALS}
Soumyadip Sengupta, Brian Curless, Ira Kemelmacher-Shlizerman, and Steven~M.
  Seitz.
\newblock A light stage on every desk.
\newblock {\em 2021 IEEE/CVF International Conference on Computer Vision
  (ICCV)}, pages 2400--2409, 2021.

\bibitem{Sevastopolsky2020Relightable3H}
Artem Sevastopolsky, SA Ignatiev, Gonzalo Ferrer, Evgeny Burnaev, and Victor~S.
  Lempitsky.
\newblock Relightable 3d head portraits from a smartphone video.
\newblock {\em ArXiv}, abs/2012.09963, 2020.

\bibitem{smith2020morphable}
William~AP Smith, Alassane Seck, Hannah Dee, Bernard Tiddeman, Joshua~B
  Tenenbaum, and Bernhard Egger.
\newblock A morphable face albedo model.
\newblock In {\em Proceedings of the IEEE/CVF Conference on Computer Vision and
  Pattern Recognition}, pages 5011--5020, 2020.

\bibitem{Stratou2011EffectOI}
Giota Stratou, Abhijeet Ghosh, Paul~E. Debevec, and Louis-Philippe Morency.
\newblock Effect of illumination on automatic expression recognition: A novel
  3d relightable facial database.
\newblock {\em Face and Gesture 2011}, pages 611--618, 2011.

\bibitem{tewari2019fml}
Ayush Tewari, Florian Bernard, Pablo Garrido, Gaurav Bharaj, Mohamed Elgharib,
  Hans-Peter Seidel, Patrick P{\'e}rez, Michael Zollhofer, and Christian
  Theobalt.
\newblock Fml: Face model learning from videos.
\newblock In {\em Proceedings of the IEEE/CVF Conference on Computer Vision and
  Pattern Recognition}, pages 10812--10822, 2019.

\bibitem{tewari2020stylerig}
Ayush Tewari, Mohamed Elgharib, Gaurav Bharaj, Florian Bernard, Hans-Peter
  Seidel, Patrick P{\'e}rez, Michael Zollhofer, and Christian Theobalt.
\newblock Stylerig: Rigging stylegan for 3d control over portrait images.
\newblock In {\em Proceedings of the IEEE/CVF Conference on Computer Vision and
  Pattern Recognition}, pages 6142--6151, 2020.

\bibitem{tewari2021learning}
Ayush Tewari, Hans-Peter Seidel, Mohamed Elgharib, Christian Theobalt, et~al.
\newblock Learning complete 3d morphable face models from images and videos.
\newblock In {\em Proceedings of the IEEE/CVF Conference on Computer Vision and
  Pattern Recognition}, pages 3361--3371, 2021.

\bibitem{tewari2018self}
Ayush Tewari, Michael Zollh{\"o}fer, Pablo Garrido, Florian Bernard, Hyeongwoo
  Kim, Patrick P{\'e}rez, and Christian Theobalt.
\newblock Self-supervised multi-level face model learning for monocular
  reconstruction at over 250 hz.
\newblock In {\em Proceedings of the IEEE conference on computer vision and
  pattern recognition}, pages 2549--2559, 2018.

\bibitem{tewari2017mofa}
Ayush Tewari, Michael Zollhofer, Hyeongwoo Kim, Pablo Garrido, Florian Bernard,
  Patrick Perez, and Christian Theobalt.
\newblock Mofa: Model-based deep convolutional face autoencoder for
  unsupervised monocular reconstruction.
\newblock In {\em Proceedings of the IEEE International Conference on Computer
  Vision Workshops}, pages 1274--1283, 2017.

\bibitem{thies2020neural}
Justus Thies, Mohamed Elgharib, Ayush Tewari, Christian Theobalt, and Matthias
  Nie{\ss}ner.
\newblock Neural voice puppetry: Audio-driven facial reenactment.
\newblock In {\em European conference on computer vision}, pages 716--731.
  Springer, 2020.

\bibitem{thies2016face2face}
Justus Thies, Michael Zollhofer, Marc Stamminger, Christian Theobalt, and
  Matthias Nie{\ss}ner.
\newblock Face2face: Real-time face capture and reenactment of rgb videos.
\newblock In {\em Proceedings of the IEEE conference on computer vision and
  pattern recognition}, pages 2387--2395, 2016.

\bibitem{torrance1967theory}
Kenneth~E Torrance and Ephraim~M Sparrow.
\newblock Theory for off-specular reflection from roughened surfaces.
\newblock {\em Josa}, 57(9):1105--1114, 1967.

\bibitem{tran2019towards}
Luan Tran, Feng Liu, and Xiaoming Liu.
\newblock Towards high-fidelity nonlinear 3d face morphable model.
\newblock In {\em Proceedings of the IEEE/CVF Conference on Computer Vision and
  Pattern Recognition}, pages 1126--1135, 2019.

\bibitem{tran2018nonlinear}
Luan Tran and Xiaoming Liu.
\newblock Nonlinear 3d face morphable model.
\newblock In {\em Proceedings of the IEEE conference on computer vision and
  pattern recognition}, pages 7346--7355, 2018.

\bibitem{Wang2022SunStagePR}
Yifan Wang, Aleksander Holynski, Xiuming Zhang, and Xuaner~Cecilia Zhang.
\newblock Sunstage: Portrait reconstruction and relighting using the sun as a
  light stage.
\newblock {\em ArXiv}, abs/2204.03648, 2022.

\bibitem{Wang2004ImageQA}
Zhou Wang, Alan~Conrad Bovik, Hamid~R. Sheikh, and Eero~P. Simoncelli.
\newblock Image quality assessment: from error visibility to structural
  similarity.
\newblock {\em IEEE Transactions on Image Processing}, 13:600--612, 2004.

\bibitem{Wang2020SingleIP}
Zhibo Wang, Xin Yu, Ming Lu, Quan Wang, Chen Qian, and Feng Xu.
\newblock Single image portrait relighting via explicit multiple reflectance
  channel modeling.
\newblock {\em ACM Transactions on Graphics (TOG)}, 39:1 -- 13, 2020.

\bibitem{weyrich2006analysis}
Tim Weyrich, Wojciech Matusik, Hanspeter Pfister, Bernd Bickel, Craig Donner,
  Chien Tu, Janet McAndless, Jinho Lee, Addy Ngan, Henrik~Wann Jensen, et~al.
\newblock Analysis of human faces using a measurement-based skin reflectance
  model.
\newblock {\em ACM Transactions on Graphics (ToG)}, 25(3):1013--1024, 2006.

\bibitem{Wu_2020_CVPR}
Shangzhe Wu, Christian Rupprecht, and Andrea Vedaldi.
\newblock Unsupervised learning of probably symmetric deformable 3d objects
  from images in the wild.
\newblock In {\em CVPR}, 2020.

\bibitem{Yamaguchi2018HighfidelityFR}
Shugo Yamaguchi, Shunsuke Saito, Koki Nagano, Yajie Zhao, Weikai Chen, Kyle
  Olszewski, Shigeo Morishima, and Hao Li.
\newblock High-fidelity facial reflectance and geometry inference from an
  unconstrained image.
\newblock {\em ACM Transactions on Graphics (TOG)}, 37:1 -- 14, 2018.

\bibitem{yang2020facescape}
Haotian Yang, Hao Zhu, Yanru Wang, Mingkai Huang, Qiu Shen, Ruigang Yang, and
  Xun Cao.
\newblock Facescape: a large-scale high quality 3d face dataset and detailed
  riggable 3d face prediction.
\newblock In {\em Proceedings of the ieee/cvf conference on computer vision and
  pattern recognition}, pages 601--610, 2020.

\bibitem{yu2021bisenet}
Changqian Yu, Changxin Gao, Jingbo Wang, Gang Yu, Chunhua Shen, and Nong Sang.
\newblock Bisenet v2: Bilateral network with guided aggregation for real-time
  semantic segmentation.
\newblock {\em International Journal of Computer Vision}, 129(11):3051--3068,
  2021.

\bibitem{Zhang2018TheUE}
Richard Zhang, Phillip Isola, Alexei~A. Efros, Eli Shechtman, and Oliver Wang.
\newblock The unreasonable effectiveness of deep features as a perceptual
  metric.
\newblock {\em 2018 IEEE/CVF Conference on Computer Vision and Pattern
  Recognition}, pages 586--595, 2018.

\bibitem{zheng2022avatar}
Yufeng Zheng, Victoria~Fern{\'a}ndez Abrevaya, Marcel~C B{\"u}hler, Xu Chen,
  Michael~J Black, and Otmar Hilliges.
\newblock Im avatar: Implicit morphable head avatars from videos.
\newblock In {\em Proceedings of the IEEE/CVF Conference on Computer Vision and
  Pattern Recognition}, pages 13545--13555, 2022.

\end{thebibliography}
}

\end{document}